\documentclass[conference]{IEEEtran}
\usepackage{times}

\usepackage[numbers]{natbib}
\usepackage{multicol}
\usepackage[bookmarks=true]{hyperref}

\usepackage{color}
\usepackage{amsfonts}
\usepackage{amsmath}
\usepackage{url}
\usepackage{cleveref}
\usepackage{comment}
\usepackage{graphicx}
\usepackage{float}
\usepackage{subcaption}
\usepackage{booktabs}
\usepackage{multirow}
\newcommand{\method}[1]{ConBaT}

\newcommand{\yue}[1]{\textcolor{orange}{Yue: #1}}

\newcommand{\rbt}[1]{\textcolor{black}{#1}}
\newcommand{\partitle}[1]{\noindent{\textbf{#1:}}}

\IEEEoverridecommandlockouts

\begin{document}

\title{ConBaT: Control Barrier Transformer\\for Safe Policy Learning}


\author{\authorblockN{Yue Meng\authorrefmark{1}\thanks{This work was done when Yue Meng was an intern at Microsoft Research.},
Sai Vemprala\authorrefmark{2},
Rogerio Bonatti\authorrefmark{2}, 
Chuchu Fan\authorrefmark{1}, and
Ashish Kapoor\authorrefmark{2}}
\authorblockA{\{mengyue, chuchu\}@mit.edu\authorrefmark{1} \quad
\{saihv, rbonatti, akapoor\}@microsoft.com\authorrefmark{2}
}
\authorblockA{\authorrefmark{1}Massachusetts Institute of Technology \quad
\authorrefmark{2}Microsoft Research
}
}

\maketitle

\begin{abstract}%
Large-scale self-supervised models have recently revolutionized our ability to perform a variety of tasks within the vision and language domains. However, using such models for autonomous systems is challenging because of safety requirements: besides executing correct actions, an autonomous agent must also avoid the high cost and potentially fatal critical mistakes. Traditionally, self-supervised training mainly focuses on imitating previously observed behaviors, and the training demonstrations carry no notion of which behaviors should be explicitly avoided. In this work, we propose Control Barrier Transformer (\method{}), an approach that learns safe behaviors from demonstrations in a self-supervised fashion. \method{} is inspired by the concept of control barrier functions in control theory and uses a causal transformer that learns to predict safe robot actions autoregressively using a critic that requires minimal safety data labeling. During deployment, we employ a lightweight online optimization to find actions that ensure future states lie within the learned safe set. We apply our approach to different simulated control tasks and show that our method results in safer control policies compared to other classical and learning-based methods such as imitation learning, reinforcement learning, and model predictive control.
\end{abstract}

\IEEEpeerreviewmaketitle

\section{Introduction}

Mobile robots are finding increasing use in complex environments through tasks such as autonomous navigation, delivery, and inspection~\cite{ning2021survey,gillula2011applications}. Any unsafe behavior such as collisions in the real world carries a tremendous amount of risk while potentially resulting in catastrophic outcomes. Hence, robots are expected to execute their actions in a safe, reliable manner while achieving the desired tasks. Yet, learning safe behaviors such as navigation comes with several challenges. 
Primarily, notions of safety are often indirect and only implicitly found in datasets, as it is customary to show examples of optimal actions (what the robot should do) as opposed to giving examples of failures (what to avoid).
In fact, defining explicit safety criteria in most real-world scenarios is a complex task and requires deep domain knowledge~\cite{gressenbuch2021predictive,braga2021semantic,kreutzmann2013towards}.
In addition, learning algorithms can struggle to directly infer safety from high-dimensional observations, as most robots do not operate with global ground truth state information.

We find examples of safe navigation using both classical and learning-based methods. 
Classical methods often rely on carefully crafted models and safety constraints expressed as optimization problems and require expensive tuning of parameters for each scenario \cite{zhou2017fast,van2008reciprocal,trautman2010unfreezing}. The challenges in translating safety definitions into rules make it challenging to deploy classical methods in complex settings. The mathematical structure of such planners can also make them prone to adversarial attacks \cite{vemprala2021adversarial}. 

Within the domain of safe learning-based approaches we see instances of reinforcement learning and imitation learning leveraging safety methods \cite{brunke2022safe,turchetta2020safe}, and also model-based learning via reachability analysis and control barrier functions \cite{herbert2021scalable,luo2022sample}.
The application of learning-based approaches to safe navigation is significantly hindered by the fact that while expert demonstrations may reveal one way to solve a certain task, they do not often reflect which types of unsafe behaviors should be avoided by the agent.
We can draw similarities and differences with other domains: natural language (NL) and vision models can learn how to generate grammatically correct text or temporally consistent future image frames by following patterns consistent with the training corpus.
However, for control tasks, notions of safety are less evident from demonstrations. While the cost of a mistake is not fatal in NL and vision, when it comes to autonomous navigation, we find that states that disobey the safety rules can have significant negative consequences for physical systems.
Within this context, our paper aims to take a step toward answering a fundamental question:
how can we use agent demonstrations to learn a policy that is both effective for the desired task and also respects safety-critical constraints?

Recently, the success of large language models~\cite{vaswani2017attention, brown2020language} has inspired the development of a class of Transformer-based models for decision-making which uses auto-regressive losses over sequences of demonstrated state and action pairs~\cite{reed2022generalist, bonatti2022pact}. While such models are able to learn task-specific policies from expert data, they lack a clear notion of safety and are unable to avoid unsafe actions explicitly. Our work builds upon this paradigm of large autoregressive Transformers applied to perception-action sequences and introduces methods to learn policies in a safety-critical fashion.

\begin{figure*}[!htbp]
\centering
\includegraphics[width=1.0\textwidth]{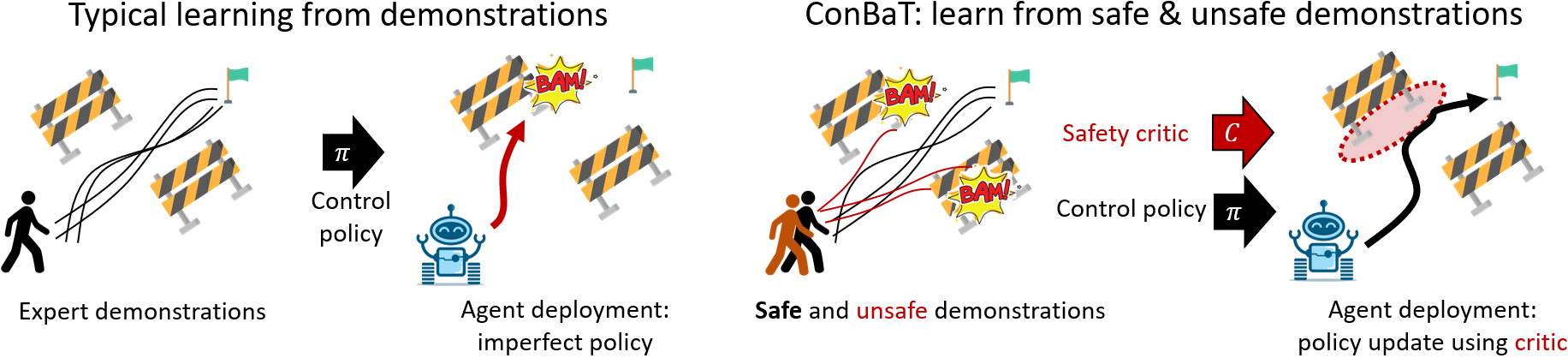}
\caption{\small (Left) An agent trained to imitate expert demonstrations may just focus on the end result of the task without explicit notions of safety. (Right) Our proposed method \method{} learns a safety critic on top of the control policy and uses the critic's control barrier to optimize the policy for safe actions actively.}
\end{figure*}

Our method, named Control Barrier Transformer (\method{}), takes inspiration in barrier functions from control theory~\cite{ames2019control}. Our architecture consists of a causal Transformer augmented with a safety critic that implicitly models a control barrier function to evaluate states for safety. Instead of relying on a complex set of hand-defined safety rules, our proposed critic only requires a binary label of whether a certain demonstration is deemed to be safe or unsafe. This control barrier critic then learns to map observations to a continuous safety score, inferring safety constraints in a self-supervised way. If a proposed action from the policy is deemed unsafe by the critic, \method{} attempts to compute a better action that ensures the safety of reachable states. A lightweight optimization scheme operates on the critic values to minimally modify the proposed action and result in a safer alternative, a process inspired by optimal control methods and enabled by the fully differentiable fabric of the model.
Unlike conventional formulations of the control barrier function, our critic operates in the embedding space of the transformer as opposed to ground truth states, making it applicable to a wide variety of systems.
We list our contributions below: 

\begin{itemize}
\item We propose the Control Barrier Transformer (\method{}) architecture, built upon causal Transformers with the addition of a differentiable safety critic inspired by control barrier functions. Our model can be trained auto-regressively with state-action pairs and can be applied to safety-critical applications such as safe navigation.
We present a loss formulation that enables the critic to map latent embeddings to a continuous safety value, and during deployment we couple the critic with a lightweight optimization scheme over possible actions to keep future states within the safe set. 
\item We apply our method to two simulated environments: a simplified F1 car simulator upon which we perform several analyses, and a simulated LiDAR-based vehicle navigation scenario. We compare our method with imitation learning and reinforcement learning baselines, as well as a classical model-predictive control method. We show that \method{} results in lower collision rates and longer safe trajectory lengths.
\item We show that novel safety definitions (beyond collision avoidance) can be quickly learned by \method{} with minimal data labeling. Using new demonstrations we can finetune the safety critic towards new constraints such as ``avoid driving in straight lines", and the adapted policy shifts towards the desired distribution.
\end{itemize}

\section{Methods}
\label{sec:methods}

We formulate the problem of learning safe control from observations as a partially observable Markov decision process that involves an agent interacting with an environment to receive high-dimensional observations $o$, which are used to take continuous-valued actions $a$. For simplicity, we treat the observations as equivalent to states $s$ while noting that actual agent states are implicit and not evident from the data. We define a trajectory $\tau$ as a set of state-action tuples $(s_t, a_t)$ over a discrete-time finite-horizon $t \in [0, T]$. 
We assume that we have access to demonstrations from two sets of trajectories: $\tau \in \Sigma_{\text{s}}$, which obey the desired safety constraints at all time steps, and $\tau \in \Sigma_{\text{u}}$, which lead to unsafe terminal states.
Our goal is to learn a safe policy $\pi_{\text{safe}}: s \rightarrow a$ that results in actions that mimic the action distribution from good demonstrations $\Sigma_{\text{s}}$, while avoiding sequences of actions that lead to the unsafe terminal states of $\Sigma_{\text{u}}$.

\subsection{Base Architecture: Perception-Action Causal Transformer}

The base transformer architecture for our method is derived from the Perception-Action Causal Transformer (PACT) proposed in \cite{bonatti2022pact}. PACT primarily focuses on creating a pretrained representation that can be finetuned towards a diverse set of tasks for a mobile agent. Its pretraining stage uses a history of state-action pairs from expert demonstrations to autoregressively train both a world model and a policy network, using imitation learning for its training objectives. 
The main functions from PACT that we employ are:
\begin{itemize}
\item \textbf{Tokenizer}: The state and action tokenizers operate on raw high dimensional observation and action data, and learn to represent them as compact tokens:  $T_s(s_t) \to s'_t$, $T_a(a_t) \to a'_t$, where $s', a' \in \mathbb{R}^{d}$ and $d$ is a fixed length of the token embedding. $T_s$ and $T_a$ are learned functions, and we detail their architectures when describing concrete problem settings in subsequent sections.
\item \textbf{Causal Transformer}: A set of Transformer blocks $X$ operates upon a sequence of state and action tokens. A causal attention mask is applied so that the transformer learns a distribution over the token at index $i$ given the history of tokens from $[0, i-1]$. The output of this module is a sequence of state and action embeddings, as $X(s'_0, a'_0, s'_1, a'_1, ... , s'_T, a'_T) \to (s^+_0, a^+_0, ..., s^+_T, a^+_T)$, where $s^+_t, a^+_t \in \mathbb{R}^{d}$.
\item \textbf{Policy model}: An action prediction head acts as a policy and operates on the output state embedding $s^+_t$ at a given timestep $t$  and predicts the appropriate action: $\pi(s^+_t) \to \hat{a}_t$.
\item \textbf{World model}: This module operates on the state and action output embeddings from the current timestep, and predicts the next state embedding: $\phi(s^+_t, a^+_t) \to \hat{s}'_{t+1}$. This module functions as a regularizing mechanism during training, and can optionally be utilized during the safe policy roll-out as explained next.
\end{itemize}

\subsection{Control Barrier Transformer}
On top of the original PACT architecture, we introduce important modifications that allow \method{} to generate safe actions.
While the original policy head does not explicitly distinguish between safe and unsafe behavior,
we instead refine action proposals using a safety critic. 
\\

\subsubsection{\textbf{Control Barrier Critic}}
\begin{figure*}
     \centering
     \includegraphics[width=0.9\textwidth]{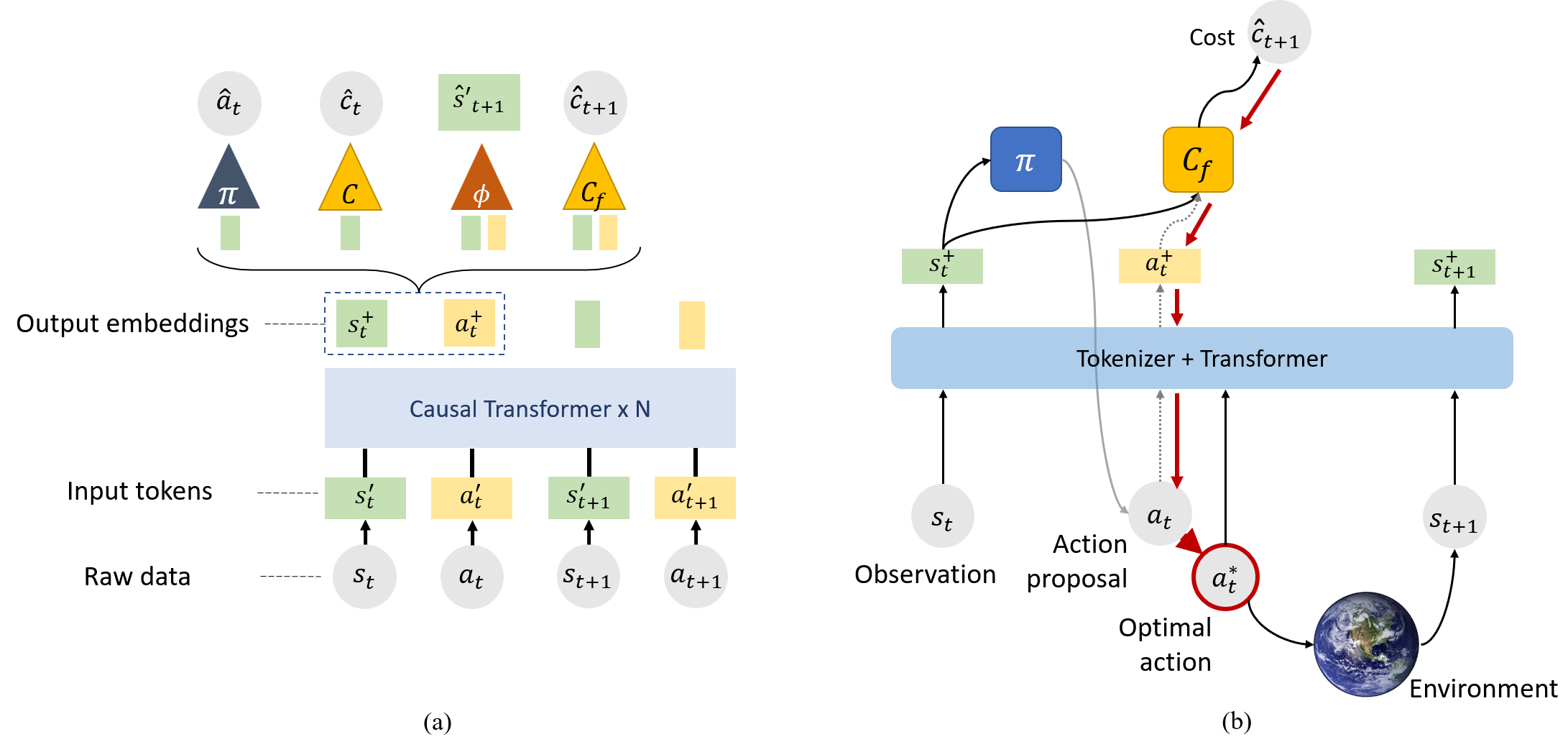}
     \caption{\small (a) The \method{} architecture - a causal Transformer operates on state and action tokens ($s', a'$) to produce embeddings ($s^+, a^+$). A policy head $\pi$ computes actions given state embeddings, and a current state critic $C$ computes a safety score. Both state and action embeddings are processed by a world model $\phi$ to compute the future state token, and by the future critic $C_f$ to produce a future safety score. (b) The deployment process for \method{} involves a feedback loop. The future critic evaluates action proposals from the policy head to check the safety of resultant states. The red arrows show the flow of gradients that allow optimizing for the safe action that results in a desired cost characteristic. The optimal action $a^*$ is used as the final command. }
     \label{fig:arch}
\end{figure*}

We augment the transformer backbone with two trainable critic modules, which predict safety scores for the current and future expected states.
The first critic $C: s^+_t \to \hat{c}_t \in \mathbb{R}$  maps the current state embedding $s^+_t$ to a real-valued safety score $\hat{c}_t$. The second critic $C_f: (s^+_t, a^+_t) \to \hat{c}_{t+1} \in \mathbb{R}$ estimates the future state safety score $\hat{c}_{t+1}$ based on the current state and action embeddings $s^+_t, a^+_t$. Here $C_f$ can be interpreted as conjoining both a world model and safety critic within a single network. 
We display our entire \method{} architecture along with the prediction heads in \Cref{fig:arch}(a).

To train these critics, we draw inspiration from the control barrier functions in classical controls literature \citep{ames2019control}. 
We assume our system to be of the form $\dot{s}=f(s, a)$ with $f$ locally Lipschitz continuous, state $s\in\mathcal{S}$ and action $a\in\mathcal{A}$. Let us denote the safe set of states as $\mathcal{S}_s\subset\mathcal{S}$. If there exists a function $h:\mathcal{S}\to \mathbb{R}$ and a policy $\pi^*:\mathcal{S}\to\mathcal{A}$ satisfying: 
\begin{gather}
h(s)\geq 0, \forall s \in \mathcal{S}_s \label{eq:cbf1} \\
h(s)<0, \forall s \in\mathcal{S}_u=\mathcal{S}/\mathcal{S}_s \label{eq:cbf2} \\
\dot{h}(s)=\partial{h(s)} /\partial s \cdot f(s, \pi^*(s)) \geq -\alpha h(s) \ \ \text{with} \ \ \alpha >0,
\label{eq:cbf3}
\end{gather}
then 
$h$ is defined to be a control barrier function (CBF), 
and the policy $\pi^*$ ensures any initial state starting from $\mathcal{S}_s$ will always stay in $\mathcal{S}_s$ (forward invariant). The proof can be found in~\citep{ames2014control}(Theorem 1). Since $\mathcal{S}_s$ denotes the safe set, the policy $\pi^*$ will guarantee all the states initialized from the safe set to be always safe.

In practice, finding the perfect safe policy $\pi^*$ for all states is challenging due to input constraints, imperfect policy learning, or disturbances~\citep{qin2021learning, dawson2022safe}. One alternative is to learn the CBF on top of an imperfect policy $\pi$ and then use an optimization routine (quadratic program~\citep{ames2016control},  second-order cone program~\citep{buch2021robust}, gradient descent) to steer $\pi$ to satisfy CBF constraints, thus achieving safety. \method{} follows a similar philosophy where the critics are learned to approximate the CBF for a given (imperfect) policy $\pi$ (from PACT) so that the CBF conditions can be satisfied in most cases, and then use back-propagation to rectify $\pi$ to satisfy the CBF condition during deployment. In our method, the supervision signal needed to learn the critic values is just the binary labels indicating a state is safe or not, 
bypassing the need for more complex hand-designed signals. We call $C$ and $C_f$ Control Barrier-like critics (CBC) and detail the learning process next.

\begin{figure}[t]
     \centering
     \includegraphics[width=0.45\textwidth]{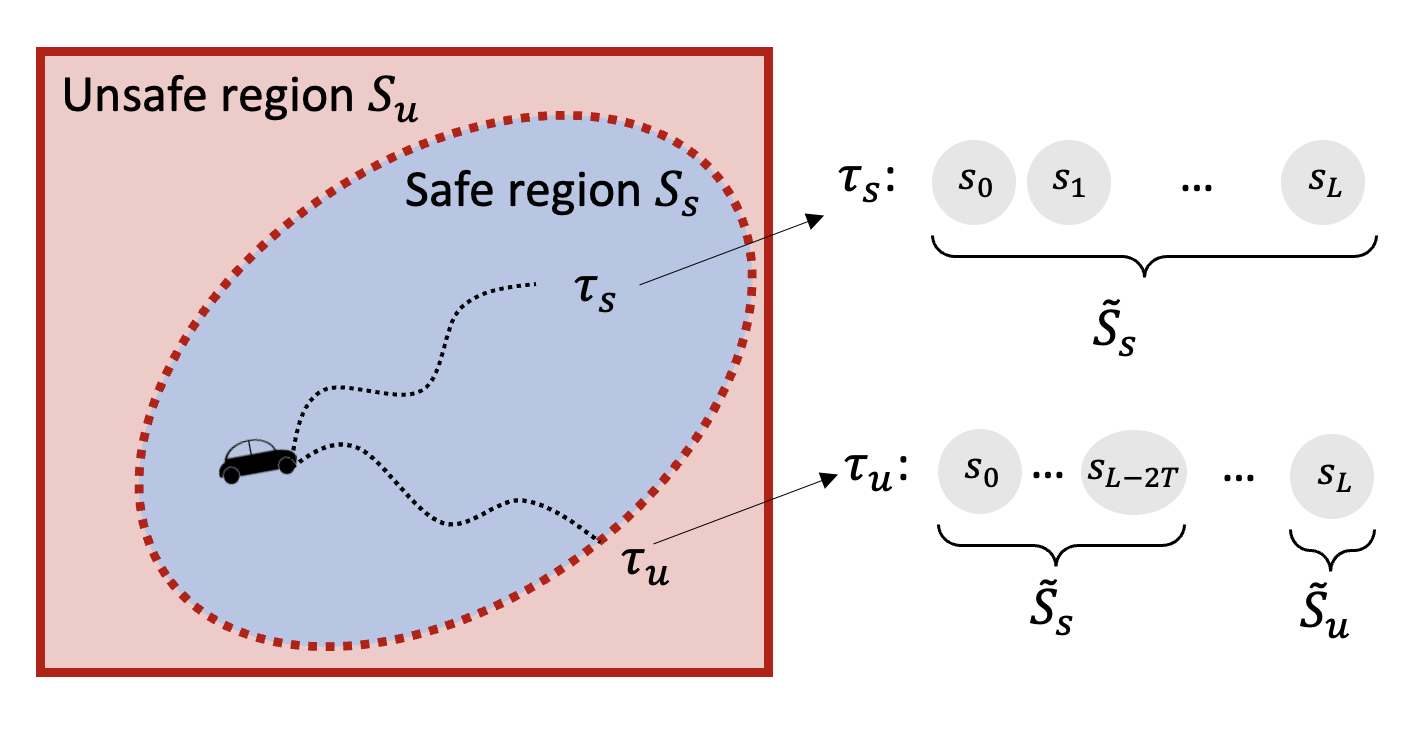}
     \caption{\small Definitions of safe and unsafe sets. In safe demonstrations $\tau_s$ all state embeddings are labeled as safe. In contrast, in unsafe trajectories $\tau_u$, only the first $(L-2T)$ embeddings are assumed to be safe, where $T$ is the Transformer context length, and only the last embedding is labeled as unsafe.}
     \label{fig:safe_regions}
\end{figure}

As mentioned earlier, $\Sigma_s$ and $\Sigma_u$ represent the safe and unsafe demonstrations. By examining the states that constitute these demonstrations, we construct a collected safe set $\tilde{\mathcal{S}}_s \in \mathcal{S}_s$ and a collected unsafe set $\tilde{\mathcal{S}}_u \in \mathcal{S}_u$. The way we achieve this is visualized in \Cref{fig:safe_regions}. For $\tilde{S}_s$, we include all the states from the trajectories in $\Sigma_s$ and the states from the first $(L-2T)$ time steps from the trajectories in $\Sigma_u$ where $L$ is the expert trajectory length, and $T$ is the time horizon that the Transformer sequence will observe. For $\tilde{S}_u$, we only consider the terminal states from the trajectories in $\Sigma_u$ (because those are the only `unsafe' states we are certain of). The embeddings corresponding to $\tilde{\mathcal{S}}_s$ and $\tilde{\mathcal{S}}_u$ can be represented as $\tilde{\mathcal{S}}_s^+$ and $\tilde{\mathcal{S}}_+$ respectively.

During CBC learning, instead of considering all possible state embeddings, we only enforce the state embeddings from the demonstrations $\Sigma_s$ and $\Sigma_u$ to satisfy the conditions above. In order for Equations (\ref{eq:cbf1})-(\ref{eq:cbf3}) to hold, we expect the critic values to be positive on the collected safe state embeddings $\tilde{\mathcal{S}}^+_s$, negative on the collected unsafe state embeddings $\tilde{\mathcal{S}}^+_u$, and to not decrease too quickly for all collected state embeddings $\tilde{S}^+=\tilde{S}_s^+\cup\tilde{S}_u^+$. 
Note that $\tilde{S}_s^+$ and $\tilde{S}_u^+$ will be highly imbalanced because not only we have more safe demonstrations than the unsafe ones but also only pick a single state from each unsafe trajectory. However, as shown empirically in future sections, we find that the critic can be learned effectively even with a highly limited number of unsafe states.

When training the CBC we use three loss terms. 
First, a classification loss $\mathcal{L}_c$ learns the safe set boundary:
\begin{equation}
    \mathcal{L}_c=\mathop{\mathbb{E}}\limits_{s^+_t\sim \tilde{\mathcal{S}}_s^+} \left[\sigma_+\left(\gamma - C(s_t^+)\right)\right] + \mathop{\mathbb{E}}\limits_{s^+_t\sim \tilde{\mathcal{S}}_u^+} \left[\sigma_+\left(\gamma+C(s_t^+)\right)\right]
\end{equation}
where $\sigma_+(x)=\max(x, 0)$ and $\gamma$ is a margin factor that ensures numerical stability in training. The second loss enforces smoothness on the CBC values over time:
\begin{equation}
    \mathcal{L}_s=\mathop{\mathbb{E}}\limits_{s^+_t\sim \tilde{\mathcal{S}}^+ }\left[\sigma_+\left((1-\alpha)C(s_t^+) - C(s_{t+1}^+) \right)\right]
\end{equation}
where $\alpha$ controls the local decay rate. Note that this loss is asymmetrical as it only penalizes fast score decays but permits instantaneous increases, as a fast-improving safety level does not pose a problem. The final loss ensures consistency between the predictions of both critics $C$ and $C_f$:
\begin{equation}
    \mathcal{L}_f=\mathop{\mathbb{E}}\limits_{s^+_t\sim \tilde{\mathcal{S}}^+}\left[\left|C_f(s_t^+, a_t^+) - C(s_{t+1}^+)\right|\right]
\end{equation}
Theoretically, one could use a single critic $C$ coupled with a world model $\phi$ to generate $\phi(s^+, a^+) \to \hat{s}'_{t+1}$ and then estimate future CBC score as $C(\hat{s}'_{t+1})$. We found it empirically helpful to use a separate critic head $C_f$ to predict future CBC scores directly from the output embeddings, as it facilitates the action optimization process described in Section~\ref{subsec:optimizer}. The total training loss is $\mathcal{L}_{CB} = \lambda_c \mathcal{L}_c + \lambda_s \mathcal{L}_s +\lambda_f \mathcal{L}_f$, with relative weights $\lambda$.
\\
\subsubsection{\textbf{Optimizing actions to improve safety}}
\label{subsec:optimizer}

As noted earlier, our goal is to learn a policy $\pi^*: s \rightarrow a$ that imitates the good demonstrations while always keeping the agent within the safe set. We compute $\pi^*$ through a two-step approach:

\partitle{Action proposal} First, we use the output of the default prediction action head $\pi$ as a reasonable action proposal $\hat{a}$. This action is then propagated through the Transformer and the future state control barrier critic, and we compute the next state's safety score: $\hat{c}_{f, t+1} = C_f(s^+_t, \hat{a}^+_t)$.

\partitle{Action optimization} If the future state is found to violate the desired safety constraint, \textit{i.e.} $\hat{c}_{f, t+1} < 0$, we need to override the default action so as to keep $s_{t+1}^+$ within the safe set $\mathcal{S}_{s}^+$. We intend to minimally modify the default action so as to result in a safe state, hence we express the optimal action as $\hat{a}_t + \Delta a^*$, and solve the following optimization problem for the modification $\Delta a$: \\
\begin{equation} 
    \Delta a^* = 
    \mathop{argmin}_{\Delta a} \lambda ||\Delta a|| + \max(-C_f(s^+_t, \hat{a}^+_t + \Delta a), 0)
\end{equation}

Given the fully differentiable model, we can efficiently compute gradients of the loss with respect to the action through the control barrier critic. 
We use a lightweight gradient descent optimization routine that results in the least action difference to the original policy to keep the agent safe. 
We then take a step in the environment, collect new observations, and repeat the process. 
We visualize the optimization step in \Cref{fig:arch}(b).

\subsection{Training procedure}

We train \method{} in a two-phase approach. 
Phase I is analogous to the original pretraining scheme for PACT \cite{bonatti2022pact}, and focuses on training the policy head, world model, tokenizers and transformer blocks.
From this phase, we aim to learn reasonable agent behavior from demonstrations.
For phase II, we freeze the base network weights and add both control barrier critic modules and train with the combination of safe and unsafe demonstrations.
By decoupling both training stages, we allow a user to potentially adapt a single base policy $\pi$ from a pretrained model towards different definitions of safety. 
Optionally, unsafe demonstrations can also be included in phase I to allow the world model to learn from the additional distribution of states, but the policy head is always trained only on safe samples. We empirically find that training the world model with data from unsafe demonstrations in the first phase results in a better final performance of the model. 
 
\section{Experimental results}
\label{sec:exp}

\begin{figure*}[!htbp]
\centering
    \begin{subfigure}[t]{0.205\textwidth}
        \includegraphics[width=\textwidth]{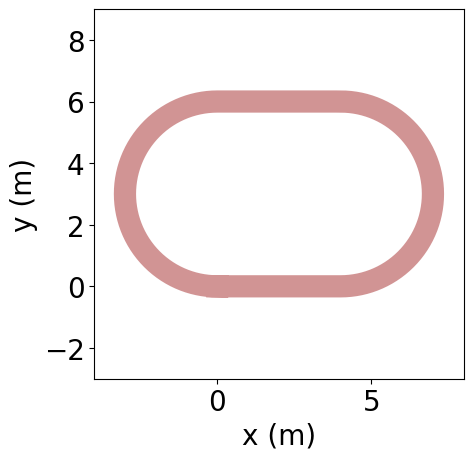}
        \caption{\small F1/10 (playground)}
    \end{subfigure}
    \quad
    \begin{subfigure}[t]{0.225\textwidth}
        \includegraphics[width=\textwidth]{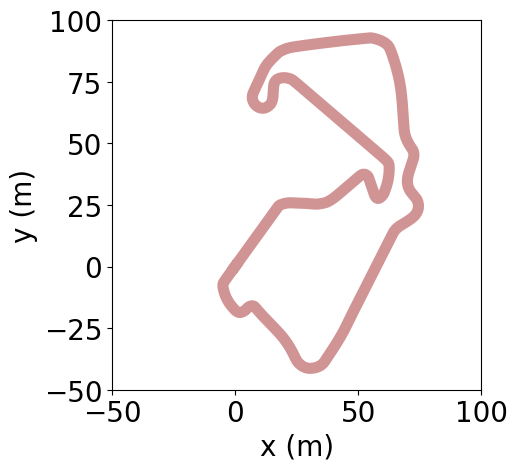}
        \caption{\small F1/10 (Silverstone)}
    \end{subfigure}
    \begin{subfigure}[t]{0.215\textwidth}
        \includegraphics[width=\textwidth]{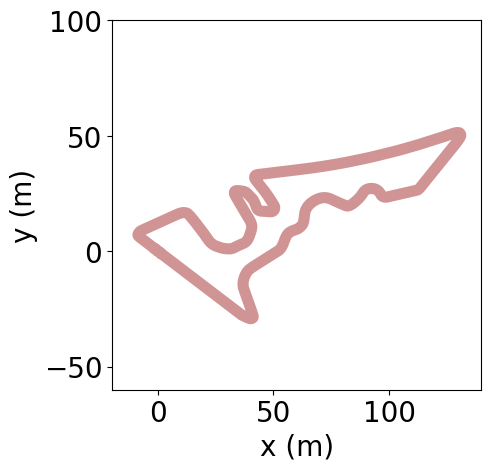}
        \caption{\small F1/10 (Austin)}
    \end{subfigure}
    \begin{subfigure}[t]{0.21\textwidth}
        \includegraphics[width=\textwidth]{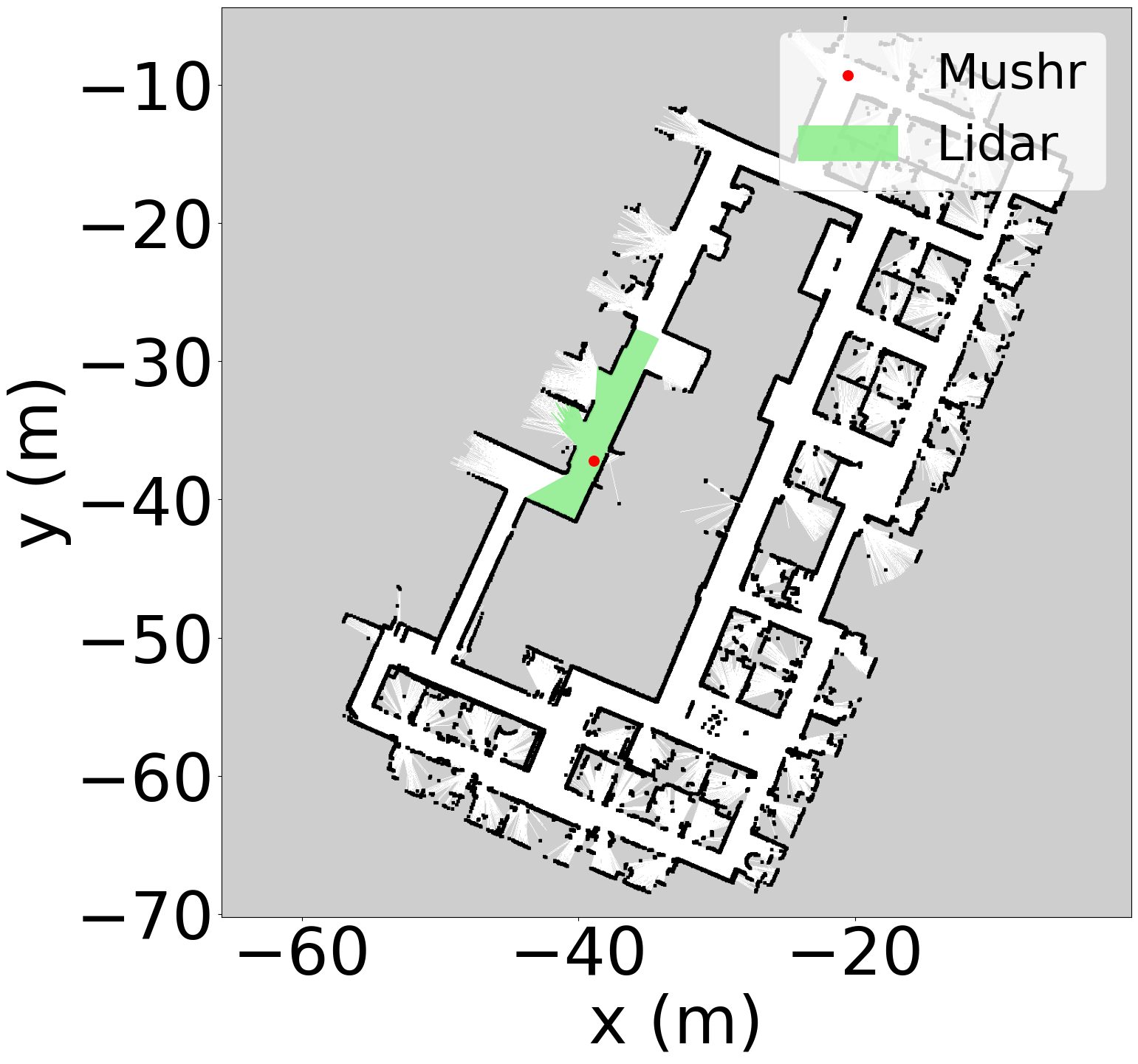}
        \caption{\small MuSHR environment}
    \end{subfigure}
    \caption{\small Simulation environment visualization.}
    \label{fig:env_viz}
\end{figure*}

We apply \method{} to two domains in simulation, which we visualize in \Cref{fig:env_viz}:\\
\textbf{F1/10 race car.} We use a simulation of a racing car that navigates within multiple 2-D racing tracks~\cite{okelly2020f1tenth} shown in \Cref{fig:env_viz}(a-c). 
The car receives observations at every step comprised of the distance and the angle relative to the center line. The action corresponds to the steering angle, and the goal is to learn how to drive safely without colliding with the track's edges. 
This is a toy scenario used to demonstrate the functionalities of our method, upon which we  conduct several ablation studies;\\
\textbf{MuSHR car.} Inspired by the MuSHR project~\cite{srinivasa2019mushr}, we use a vehicle simulation in a more realistic navigation setting. 
The car is equipped with a 2D range sensor, returns LiDAR scans as observations, and takes steering angles as actions. 
We use a 2D map scanned from a real office space of approximately $30\times70$ meters of area, for both data collection and safe navigation deployment without collisions. We show a figure of this map in \Cref{fig:env_viz}(d).

We collect sets of safe and unsafe trajectories in both domains. Each trajectory has a discrete binary safety label, which is used to train the safety critic. 
Additional details on the domains and data collection can be found in \Cref{sec:env}. 

We evaluate our models in both domains by deploying the learned policies and evaluating the rollout trajectories. 
We measure the following metrics: (1) \emph{collision rate}, measured as a percentage of trajectories in the test set that end in a crash within the cut-off time horizon; and (2) \emph{average trajectory length} (ATL), which corresponds to the average length of deployment trajectories, expressed in number of time steps before crashing or time-out if no crash occurs.

\subsection{Safe Navigation Analysis}

We first examine the safe navigation performance of \method{} in the F1/10 simulator, and compare it with PACT. We train the models with 1K demonstrations, each 100 timesteps long, from the \textit{Playground} track (\Cref{fig:env_viz}(a)). 
During deployment, we roll out 128 trajectories for each model for a maximum of 1000 timesteps. \method{} achieves a 0\% collision rate whereas PACT collides at some point in every instance.

\begin{figure}
    \centering
    \includegraphics[width=0.495\textwidth]{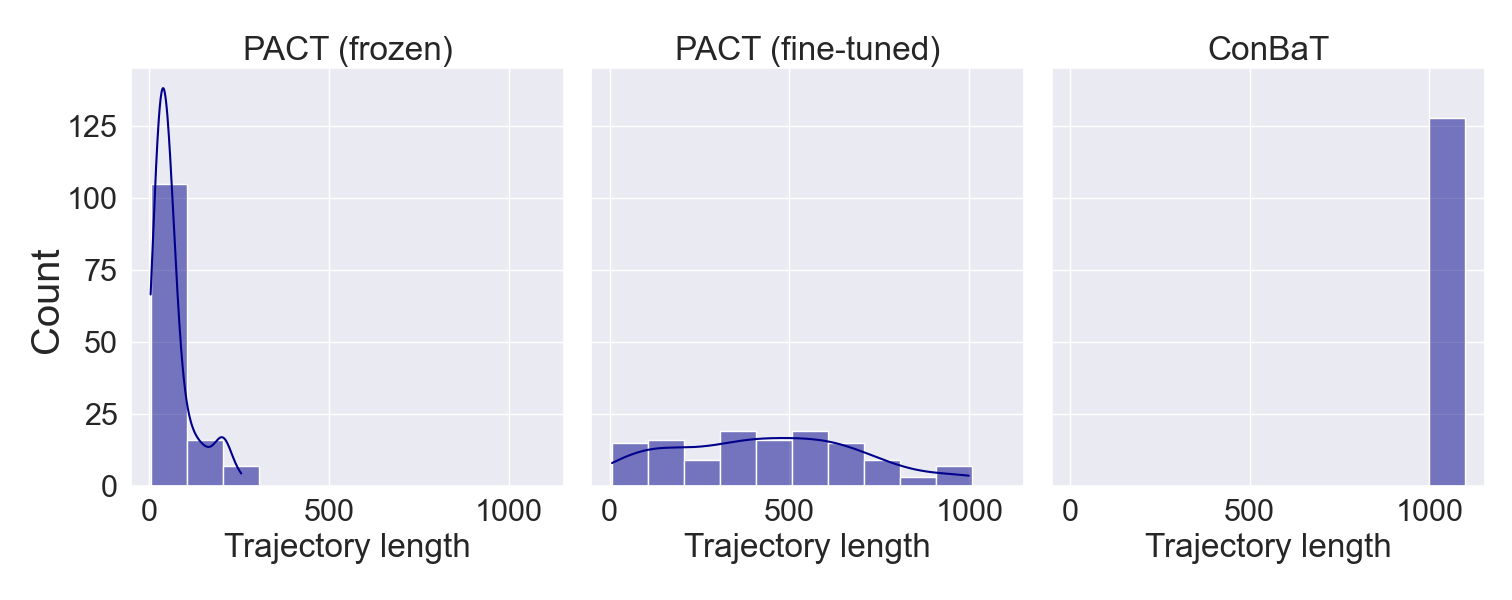}
    \caption{\small Histogram of trajectory lengths in the Silverstone track: \method{} generalizes efficiently to a new track, achieving a 100\% success rate, whereas both frozen and finetuned PACT fall short.}
    \label{fig:f1_hist}
\end{figure}

Next, we apply the \method{} model trained only on \textit{Playground} on more challenging tracks such as \textit{Silverstone} and \textit{Austin} (\Cref{fig:env_viz}(b),(c)).
We compare our model's performance against a frozen PACT model (PACT - trained only on \textit{Playground}) as well as against a version finetuned on these tracks (PACT-FT).
In \Cref{tab:f1_perf} we observe that \method{} is able to learn the safety concepts effectively enough to generalize to new tracks, outperforming both PACT and PACT-FT. 
In \Cref{fig:f1_hist}, we take a closer look at the performance on \textit{Silverstone} and show a histogram of the trajectory lengths across the test dataset. 
A PACT model trained on \textit{Playground} registers fairly low ATL values, whereas PACT-FT 
performs better by using more training samples, sometimes even reaching the maximum length of 1000. 
However \method{} significantly outperforms both, with all the trajectories being collision-free.

\begin{table}[!htbp]
\footnotesize
\centering
\caption{\small Comparison of PACT and \method{} for the F1/10 task.}
\begin{subtable}{0.495\textwidth}
    \centering
    \caption{Collision Rate (\%) - lower is better}
    \begin{tabular}{cccc}
    \toprule
                & PACT & PACT-FT & ConBaT        \\ \midrule
    Playground  & 100  & - & \textbf{0.0}  \\
    Silverstone & 100  & 96.88 &\textbf{0.0}  \\
    Austin        & 100  & 100 & \textbf{61.7} \\
    \bottomrule
    \end{tabular}
\end{subtable}
\qquad
\begin{subtable}{0.495\textwidth}
    \footnotesize
    \centering
    \caption{Avg. Trajectory Length - higher is better}  
    \begin{tabular}{cccc}
    \toprule
                & PACT   & PACT-FT & ConBaT          \\ \midrule
    Playground  & 175.45 & - & \textbf{1000} \\
    Silverstone & 61.57  & 439.28 & \textbf{1000} \\
    Austin        & 57.11 & 165.12 & \textbf{678.14} \\
    \bottomrule
    \end{tabular}
\end{subtable}
\label{tab:f1_perf}
\end{table}

\begin{figure*}[!htbp]
    \centering
    \includegraphics[width=0.95\textwidth]{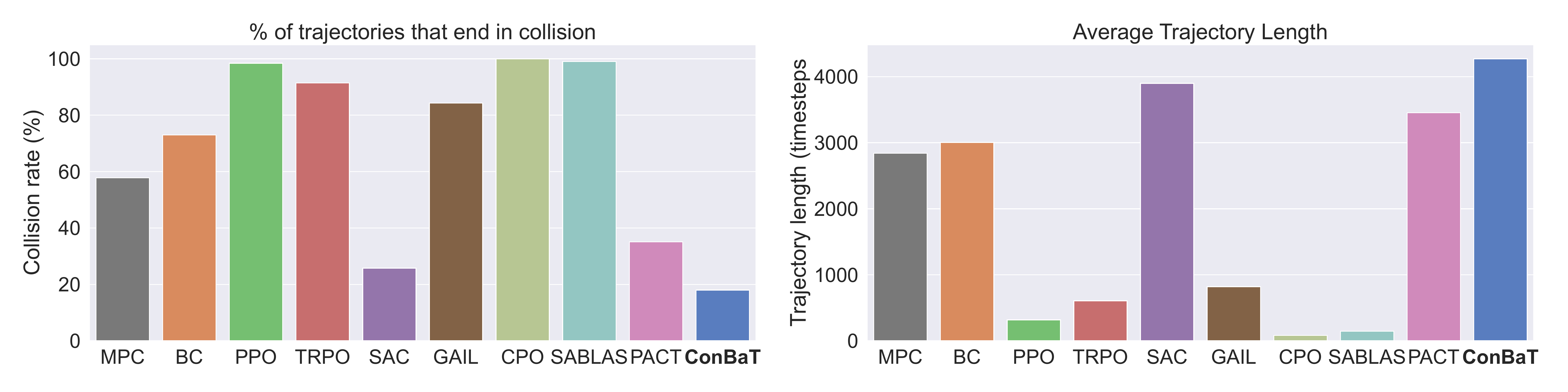}
    \caption{\small \method{} outperforms classical MPC and several learning-based methods on safe navigation in the 2D MuSHR car domain.}
    \label{fig:mushr_baselines}
\end{figure*}
\begin{figure}
\centering
    \begin{subfigure}[t]{0.2\textwidth}
        \includegraphics[width=\textwidth]{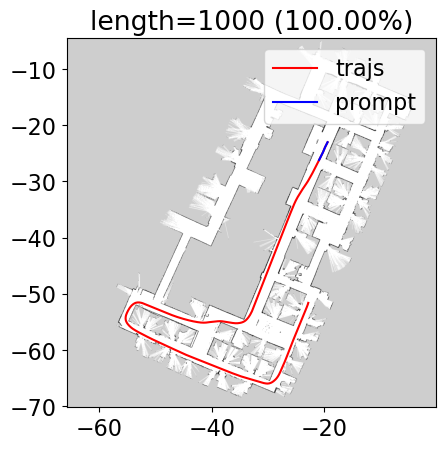}
        \caption{\small ConBaT Trajectory}
    \end{subfigure}
    \quad
    \begin{subfigure}[t]{0.2\textwidth}
        \includegraphics[width=\textwidth]{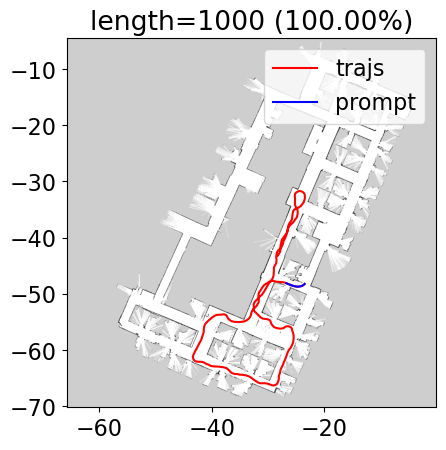}
        \caption{\small ConBaT-NS Trajectory}
    \end{subfigure}
    \hfill
    \begin{subfigure}[t]{0.4\textwidth}
        \includegraphics[width=\textwidth]{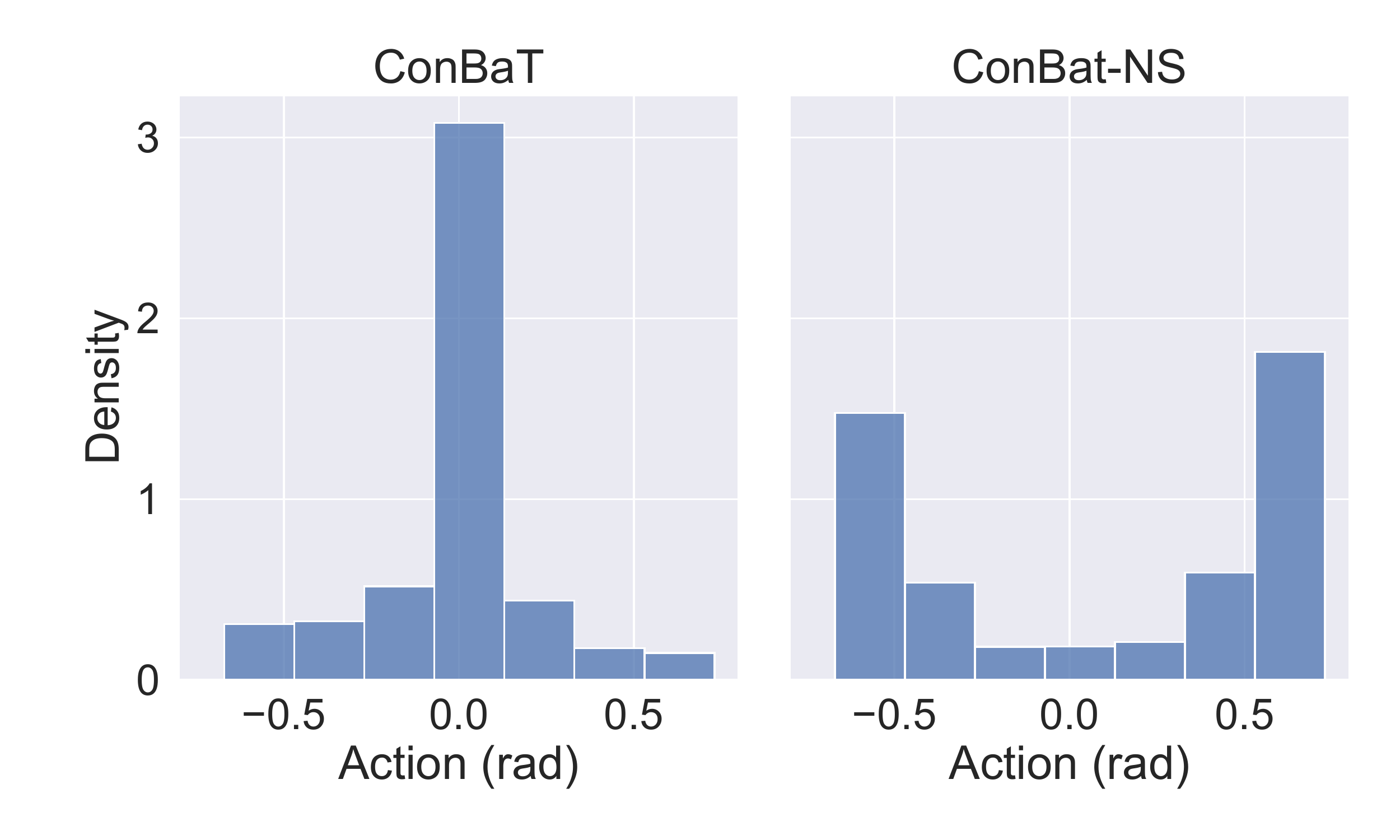}
        \caption{\small Comparison of action distributions}
        \label{fig:new_concept_hist}
    \end{subfigure}
    \caption{\small Demonstration of ConBaT being used for a new safety constraint where actions mapping to the vehicle going straight are deemed unsafe.}
    \label{fig:new_concept}
\end{figure}

\subsection{Comparison with Baseline Methods}

Using the MuSHR simulation we collect 10K trajectories to create a dataset with safe and unsafe trajectories, as described in \Cref{sec:env}. 
We then train different baseline safety models for comparison against \method{} using three different classes of algorithms: 

\begin{itemize}
\item \textit{Imitation learning (IL)}: Naive behavior cloning (BC), PACT, and Generative Adversarial Imitation Learning (GAIL) \cite{ho2016generative}; 
\item \textit{Reinforcement learning (RL)}: PPO~\cite{schulman2017proximal}, TRPO~\cite{schulman2015trust}, and SAC~\cite{haarnoja2018soft}. 
\item \textit{Classical control}: We implement a model predictive control baseline via the CasADi solver~\cite{Andersson2018}.
\item \textit{Safe control}: Constrained policy optimization (CPO)~\citep{achiam2017constrained} and a model-based safe control approach SABLAS~\citep{qin2022sablas}.
\end{itemize}

The motivation for our choice of a diverse set of baseline algorithms stems from the perspective of a user trying to solve the problem of safe navigation, who would likely select one of these these established approaches found in the robotics literature.
Across these algorithms it is evident that the underlying signals for each class such as RL/IL, and our proposed method ConBaT are different. However, our ablations intend to help the end user wishes to deploy the safest and most effective robot policy.
Further details on their implementations can be found in \Cref{sec:impl}. 
During simulation deployment, we run 128 trajectories for each model, each for a maximum of 5000 timesteps. 
\Cref{fig:mushr_baselines} shows \method{} compared against all baselines. 
Even on this complex setting using high-dimensional LiDAR observations in a realistic map, we observe that \method{} achieves the lowest collision rate and the highest ATL. 
The inferior result of CPO and SABLAS could result from the extremely high dimension system dynamics (720x2=1440) and the high dimension constraint function, whereas our approach does not need to learn the explicit system dynamics or learn from the constraint on the explicit state space (we learn the implicit system dynamics and safety concept from the safe/unsafe demonstrations).
\begin{figure*}[tbp]
\centering
\begin{subfigure}{0.4\textwidth}
\centering
\includegraphics[width=0.99\textwidth]{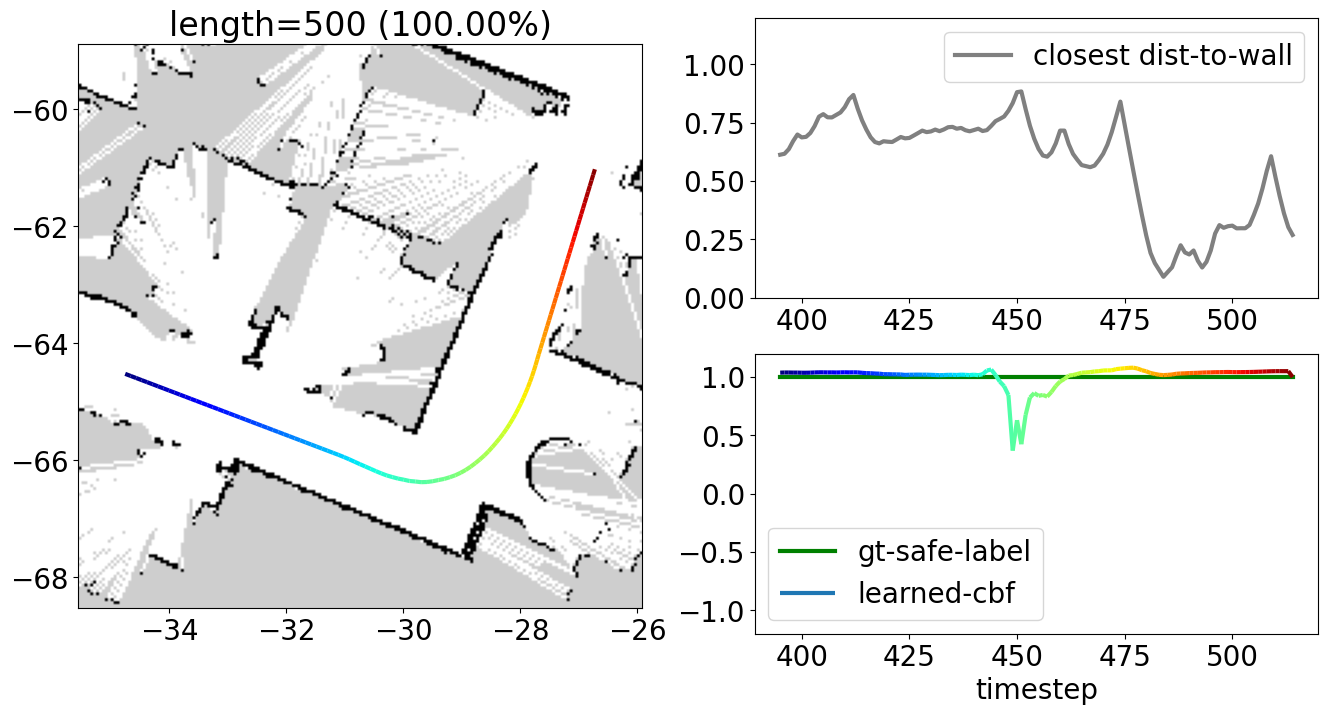}
    \caption{Safe case: CBF value is always positive. The CBF also quantifies the difficulty to escape from a potential future collision (timestep 450).}
    \label{fig:sim_cbf_viz_main_1}
\end{subfigure}
\quad
\begin{subfigure}{0.42\textwidth}
\centering
\includegraphics[width=0.99\textwidth]{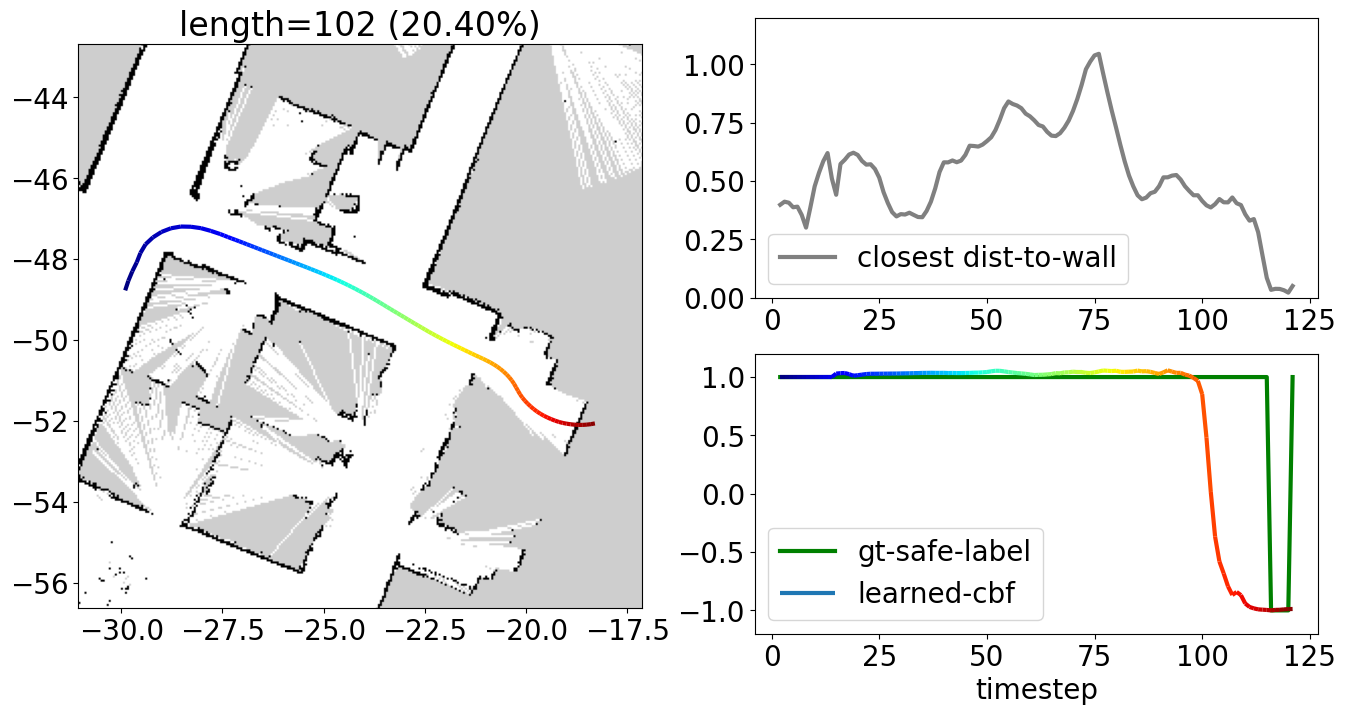}
    \caption{Unsafe case: CBF value decreases towards negative value as the agent approaches irrecoverable states, eventually leading to collision (timesteps 120 to 125).}
    \label{fig:sim_cbf_viz_main_2}
\end{subfigure}
\caption{\rbt{Qualitative assessment of CBF values in MuSHR car trajectories. The trajectory color indicates different time steps.}}
\label{fig:sim_cbf_viz_summary_main}
\end{figure*}

\subsection{Learning a new safety definition}

An important feature for a safety critic is to be able to learn different definitions of safety without major architectural changes or hand-crafting cost functions.
Therefore we evaluate \method{}'s ability to learn new safety constraints beyond simple notions of collision-free navigation. 
We consider the hypothetical case of a vehicle agent where traveling in a straight line is undesirable for the end user, and only curved motions are allowed (due to dynamical constraints, or perhaps to create a dizzy rider experience in an amusement ride). 
We generate a new dataset with unsafe labels for every straight trajectory segment, and train a new critic which we call \method{}-NS (not-straight). Note that we don't retrain or finetune the policy head, the world model, or the backbone on this new dataset. All we expect is to have the updated critic to guide the policy head to choose the ``safe" behaviors.

As seen in \Cref{fig:new_concept}, \method{}-NS succeeds in moving the vehicle through curved trajectories only, even in straight hallways, showing that new constraints can be incorporated with ease. 
\Cref{fig:new_concept_hist} also displays how this new constraint is reflected in the policy's action distribution output.
We note that both \method{} and \method{}-NS are finetuned over the same base PACT model, which demonstrates the potential for training multiple critics mapping to distinct safety constraints.

\subsection{CBF Visualization}
We provide a qualitative assessment of our learned control barrier critic along typical trajectory sequences in the MuSHR simulation (Fig.~\ref{fig:sim_cbf_viz_summary_main}).
There we can verify that the CBF value is positive for safe states and negative for unsafe vehicle states.
In addition, the smoothness loss function used during the critic's training enforces a decreasing trend of the CBF as the agent approaches a potential collision in the near future (timesteps 100 until 120 in Fig.~\ref{fig:sim_cbf_viz_main_2}).

\subsection{Ablation studies}
\label{sec:abl}
\partitle{Variations in critic architecture}

As discussed in \Cref{sec:methods}, we can use different network structures to function as the control barrier critic.
We perform experiments to compare the following architectures:\\
\emph{CBC-NW}: Uses both $C$ and $C_f$, but no world model $\phi$;\\
\emph{CBC-CW}: Only uses the current state critic $C$, and the output of the world model is fed into it for the next state's cost; \\
\emph{CBC-TF}: Uses $C$, $C_f$, and $\phi$. The input to $C_f$ is the output state embedding and the action input token $(s^+_t, a'_t)$. \\
\emph{CBC-EF}: Uses $C$, $C_f$, and $\phi$. The input to $C_f$ is the output state embedding and the output action embedding $(s^+_t, a^+_t)$.

We compare all methods in Table~\ref{tab:critic_arch}. 
We note that the world model is indeed useful, as evidenced by the lowest performance of CBC-NW. 
Using a future state critic results in slightly better performance than only using a current state critic that takes world model simulated states, but higher computational effort.
Lastly, CBC-EF achieves higher performance than CBC-TF, indicating that the critic benefits from the context that plays a part in the computation of the action embedding, as opposed to the single token. 
We note that CBC-EF is used for all other \method{} experiments in the paper.

\begin{table}[!htbp]
\centering
    \caption{\small{Comparison of architectures, F1 simulator}
    }
    \resizebox{0.4\textwidth}{!}{%
    \begin{tabular}{cccc}
    \toprule
                 Arch. & Collision (\%) & ATL (\# steps) & Runtime (s)      \\ \midrule
    PACT   & 100	&      175.45	& 63.328   \\
    CBC-NW  & 4.69	& 952.47 &	129.595  \\
    CBC-CW  & 1.56&	983.48&	200.153      \\
    CBC-TF  & 3.91	&972.42	&78.159      \\
    CBC-EF  & 0.00	& 1000	&130.032     \\
    \bottomrule
    \end{tabular} 
    }
    \label{tab:critic_arch}
\end{table}

\partitle{Data Requirements}
\label{sec:abl-data}

\begin{figure}
\centering
\includegraphics[width=0.4\textwidth]{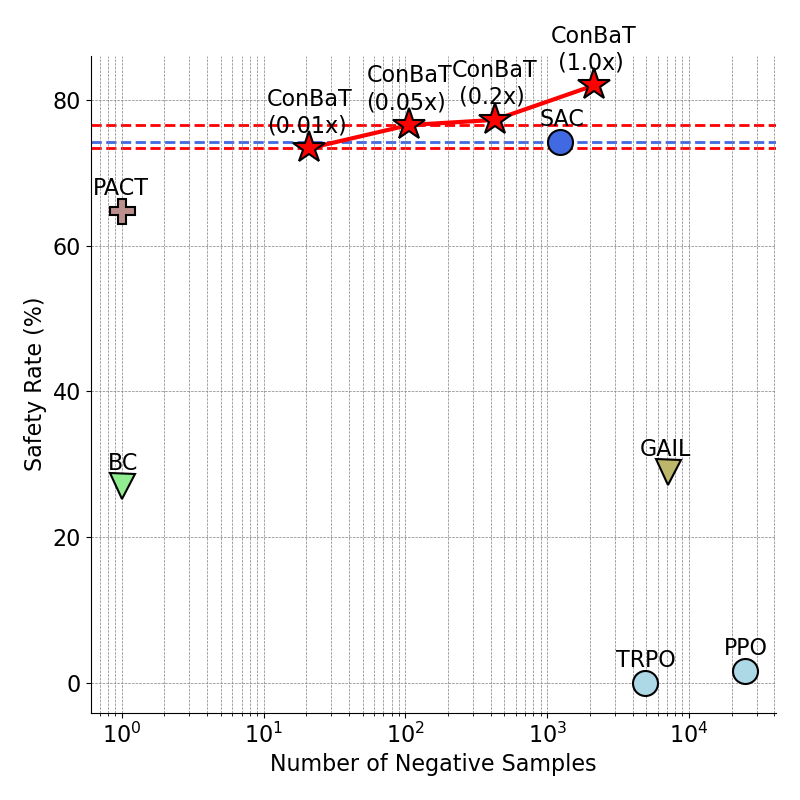}
\caption{\small{Roll-out performance of policies under different numbers of negative samples (terminal crashes) observed in the training phase. \method{} is the most efficient method in leveraging negative experiences: with only 21 negative examples, it can already learn the safety concept and improve the safety rate of based model (PACT) by 9\%, and outperforms almost all the baselines.}}
\label{fig:samp_perf_viz}
\end{figure}

The existence of demonstrations that include unsafe terminal states is critical for \method{} to be able to encode the safety constraints. 
Yet, given that simulating and collecting a large amount of unsafe demonstrations can be challenging (and costly in real world conditions), we perform a analysis of how \method{} performs in low negative data regimes when training the critic (phase II).
\Cref{tab:data_reqs} reports performance levels for varying unsafe dataset sizes.
We observe that \method{} is effective at achieving reasonable safe navigation even with small amounts of unsafe demonstrations.

We perform a more extensive analysis of data requirements in the MuSHR domain. We compare the performance of RL and IL baselines with \method{} along with how many unsafe trials were utilized in the training process. We evaluate the percentage of safe trajectories executed across 128 initial conditions with each trajectory capped at a maximum of 5000 timesteps. For \method{}, we use different sizes of unsafe trials (ranging from 21 examples, which is 1\% of the full unsafe dataset to 2131, which is 100\%). As shown in Fig.~\ref{fig:samp_perf_viz}, even just using 21 negative samples in training, the success rate of \method{} already outperforms almost all the baselines except SAC (which is 1\% higher). By leveraging just a few more (106) negative examples, we already surpass all the baselines considered in this paper. This again showcases the data-efficiency of our algorithm, contrasting it with the high number of unsafe trials usually required by RL.

\begin{table}[!htbp]
\centering
\caption{\small{Amount of unsafe trajectories used in phase II training}
    } 
    \centering
    \resizebox{0.4\textwidth}{!}{%
    \begin{tabular}{cccc}
    \toprule
               \% Trajs & Collision (\%) & ATL (\# steps) & Runtime (s)        \\ \midrule
     0  & 100 & 	175.45	& 63.328
 \\
    5  & 5.47& 	944.67	& 120.427
 \\
    10 & 3.91&	960.18	& 132.079
 \\
    25  & 1.56&	983.47	&130.049
 \\
    50   & 3.91&	960.26	&127.275

 \\
    100   & 0.78	&991.22&	127.073
\\
    \bottomrule
    \end{tabular}
    }
       
    \label{tab:data_reqs}
\end{table}

\partitle{Additional experiments in appendix}
\label{sec:abl-additional}

Besides complete implementation and architectural details, we offer additional ablation experiments for interested readers in the appendix section. 
For example, in Table~\ref{tab:effort} we compare computational efforts for all baseline methods, in Tables~\ref{tab:optim_ablations} and \ref{tab:abl_smoothness} we compare the effect of different training hyper-parameters in the policy results. In Table~\ref{tab:abl_arch} we study how different architecture backbones affect policy performance.
In addition, in Fig.~\ref{fig:optimality_test} we compare the trade-off between safety and optimality for a goal-reaching task in MuSHR using PACT and \method{}.

\section{Background}
\partitle{Learning Policies from Data}
Imitation Learning (IL) is a popular method for policy learning, and requires expert demonstrations ~\cite{schaal1996learning,atkeson1997robot,pastor2009learning,behbahani2019learning,billard2013robot,ekvall2008robot}.
The most naive IL approach is behavioral cloning (BC)~\cite{bain1995framework}, which directly learns the mapping from states to actions using supervised learning from expert data. 
Such simple applications of IL cannot generalize to the out-of-distribution scenarios induced by on-policy deployment, and often requires additional training routines~\cite{ross2011reduction}. 
Inverse reinforcement learning (IRL)~\cite{ng2000algorithms, natarajan2010multi} can mitigate this challenge, as it seeks to recover the original expert's cost function.
Alternatively, one may use generative adversarial imitation learning (GAIL)~\cite{ho2016generative, bhattacharyya2018multi}, which employs a discriminator to differentiate an expert's policy from other behaviors generated by a policy network. 
When it comes to safety, however, none of the above methods incorporate the concept of safety directly into their learning procedures. 

Reinforcement learning can enable safe policy learning through rewards that not only encourage good behaviors but also penalize unsafe states. \cite{bhattacharyya2019simulating} use an augmented reward for traffic simulation to guide safe imitation, and we find numerous examples of safe policy design~\cite{liu2020robust,berkenkamp2017safe,cheng2019end,li2019temporal}. 
However, manual reward shaping is laborious and nontrivial, often leading to undesired consequences.
In comparison, our approach with learned control barrier functions uses minimal safety labels.

\partitle{Safe policy learning using control barrier functions}
Control barrier function (CBF) is a classical concept to ensure system safety~\cite{prajna2004safety,wieland2007constructive,ames2014control,ames2019control}. Recently, several works have demonstrated learning safe policies based on existing CBFs~\cite{chen2017obstacle,borrmann2015control}, constructing CBF jointly with the policy learning via Sum-of-Squares
~\cite{wang2018permissive}, leveraging SVMs~\cite{srinivasan2020synthesis} or Neural Networks~\cite{ferlez2020shieldnn,qin2021learning,meng2021reactive,dawson2022learning,dawson2022safe}. However, these methods often assume access to raw state inputs and ground truth safety labels from the environment. Instead, \method{} works on the embedding space akin to models that plan using imagination
~\cite{okada2021dreaming,wu2022daydreamer}, learns purely from offline data, and employs an online policy rectification process to achieve safety.

\partitle{Safe policy learning with predictive world models}
\method{} is aligned with the Mode-2 proposal in~\cite{lecun2022path}, where the agent makes action proposals, evaluates the costs from future predictions, and then plans for the next action. Predicting the next state based on the current state and action is fundamental to model predictive control~\cite{bryson2018applied}. The idea of predicting the future costs via the world model can be traced back to ~\cite{schmidhuber1990making}. When it comes to high-dimension state/observation space, latent embedding dynamics are learned to help the agent to achieve high performance in reinforcement learning~\cite{okada2021dreaming,wu2022daydreamer}, and sequence predictions~\cite{giuliari2021transformer,chen2021decision,micheli2022transformers}. In our case, we learn a predictive critic on the embedding space with CBF conditions as guidance.

\section{Conclusions}

In this work, we propose \method{}, a framework that learns safe and effective navigation policies directly from positive and negative demonstrations. Our method leverages causal Transformers coupled with a differentiable safety critic that is inspired by control barrier functions in control theory. The control barrier critic module implicitly builds a safe set for states from discrete safety labels, bypassing the need for complex mathematical formulations of safety constraints. We apply our method to two simulated domains and show that our method outperforms existing classical and learning-based safe navigation approaches. Furthermore, we show that \method{} can be quickly adapted to new safety constraints from limited demonstrations. This paradigm of learning safety constraints implicitly from minimal labels reduces training effort in creating safe agents and makes it easier to adapt to new definitions of safety.

Though empirically \method{} can learn safer policies and quickly adapts to new safety concepts, there are some limitations. Our method might not guarantee safety if \method{} makes wrong CBF score predictions due to out-of-distribution data, or if the online action optimization falls into a local minimum. One solution is to follow a paradigm like DAGGER~\cite{ross2011reduction} to keep exploring the environment and collect new states and labels to update our CBF critics and world model. The runtime for \method{} in evaluation is also $0.2\sim 1.0$x higher than other learning-based methods due to the back-propagation step in online optimization. We discuss different  \method{} architectures in Sec.~\ref{sec:abl} to weigh the trade-off between the safety performance and the runtime. Further runtime improvement can be made by faster back-propagation algorithms optimized for GPU hardware. 
Currently, our framework was only applied in simulated environments. In the future, we wish to extend our approach to image-based control in 3D and toward more complex robotic platforms in simulation and the real world.
\section{Reproducibility Statement}
We are working towards providing an open-sourced implementation of our datasets, simulators, models and evaluation framework in the near future. We will make the links available for the camera-ready version of this work.

\bibliographystyle{unsrtnat}
\bibliography{z7_reference.bib}

\clearpage
\appendix
\section{Environments} 
\label{sec:env}

\subsection{F1/10 Race Car}

This simulator contains an agent that moves along a two-dimensional racetrack, which is modeled after well known F1 tracks downscaled to 1:10, as used in \cite{okelly2020f1tenth}. The racetrack is assumed to be 2 m wide, and the observation space for the agent is two dimensional, reporting the distance to the center line, as well as the relative angle from it. At every step, the agent takes an action $a \in [-1, 1] \text{rad}$ which indicates the steering angle. We use three tracks in our experiments: \textit{Playground}, \textit{Silverstone}, and \textit{Austin}, which can be visualized through \Cref{fig:env_viz}(a)-(c). 

To generate expert trajectory data from this environment, we create an expert planner using search-based model predictive control (MPC) which is able to generate collision-free paths between randomly sampled start and goal states within the track. \rbt{To delibrately produce unsafe demonstrations, we randomly decrease the safety threshold in the MPC planner to generate trajectories that will crash. We generate 1k trajectories with 711 safe trajectories and 289 unsafe trajectories. We randomly sample 800 trajectories for training, and use the rest 200 trajectories for evaluation. The average length of the collected trajectories is 100.}

\subsection{MuSHR car simulator}
MuSHR is a robot car equipped with a 2-D LiDAR sensor. The LiDAR sensor scans the environment around the car using 720 laser beams (with an angular resolution of $0.5\deg$) and returns an observation of shape $[720, 2]$, where each element is the x,y coordinate to the closest surface for that ray angle. Similar to before, the MuSHR car also takes a steering angle as the input action, which is of the range $a \in [-0.34, 0.34]$ rad in the expert demonstrations. 

\begin{figure}[!htbp]
\centering
    \includegraphics[width=0.45\textwidth]{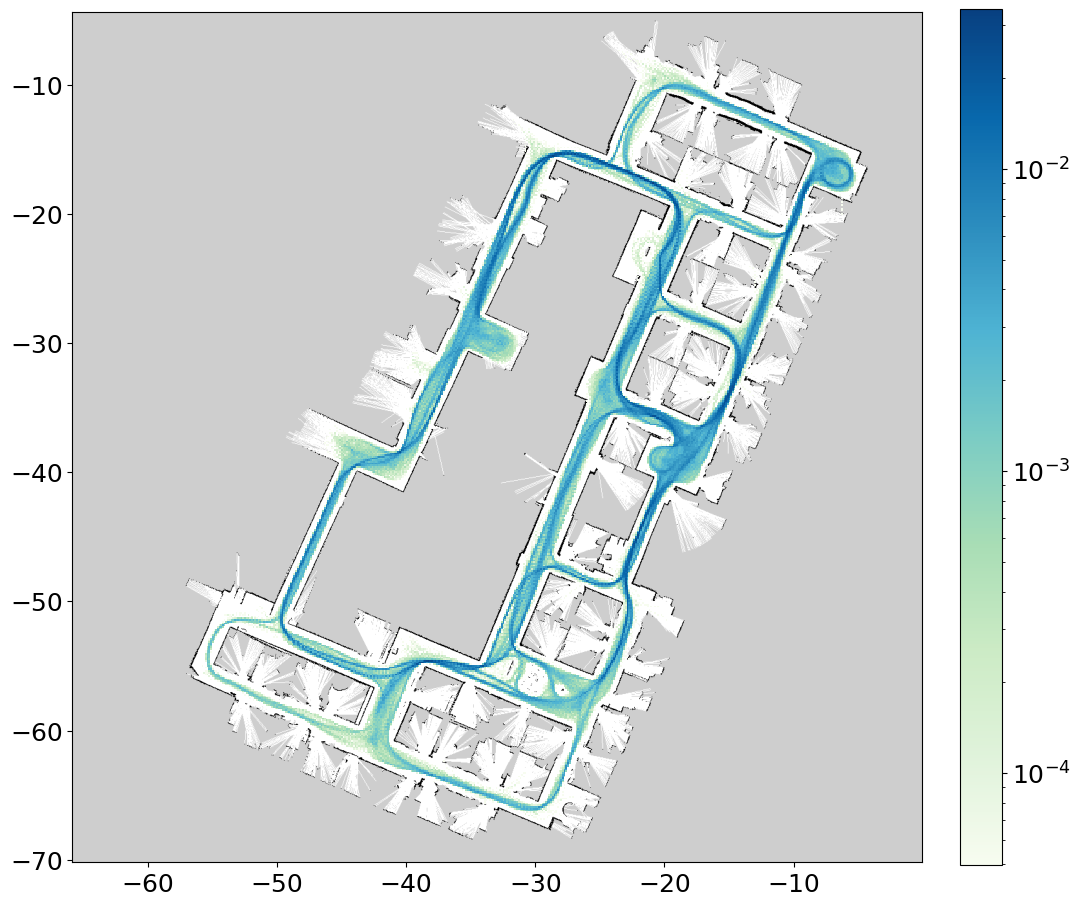}
    \caption{\rbt{Visualization of the MuSHR environment occupancy map and collected data distribution}}
    \label{fig:mushr_stat}
\end{figure}

We create a simulator for this vehicle which takes a 2D occupancy map as an input, and instantiates the vehicle dynamics and the sensor model within it. We use a pre-mapped 2D office environment for the simulation which can be visualized in \Cref{fig:env_viz}(d). We build a probabilistic roadmap over the environment and sample start and goal states in the free space, from which we generate 10K trajectories in total, where each trajectory spans around 110 timesteps. Similar to above, we use a search-based MPC for computing the collision-free trajectories. We apply slight perturbations to the expert planner to also occasionally result in unsafe trajectories. \rbt{We generate 7550 safe demonstrations and 2450 unsafe demonstrations. However, as shown in Supplementary C.7, \method{} can be trained with just 0.01X of the unsafe demonstrations (21 trajectories) here to outperform almost all the baselines. A heatmap showing the coverage of the map in the collected trajectories is shown in Figure~\ref{fig:mushr_stat}}

\section{Implementation}
\label{sec:impl}

\subsection{\method{} Implementation}

\partitle{Transformer and Control Barrier Critic} \\
For both the F1/10 simulator and for MuSHR, we use a linear  embedding to convert the observation and action into state and action tokens respectively.
For F1/10, we train the policy and world model in phase 1 for 10 epochs, and then further train the control barrier critic for 10 epochs. For the MuSHR data, we train the policy for 50 epochs and further train the control barrier critic for 10 epochs. We use the Adam optimizer for training. We list the hyperparameters used for model training in \Cref{tab:hparams}.

\begin{table}[ht]
\footnotesize
\centering
\caption{Hyperparameters}
\begin{tabular}{ll}
\toprule
Hyperparameter        & Value \\
\midrule
\# of layers          &   2    \\
\# of attention heads &   8    \\
Embedding length      &   64    \\
Sequence length       &   16    \\
Batch size            &   32    \\
Learning rate         &   1e-4    \\
Optimizer             &   Adam      \\
\# of cbf layers      &   2        \\
\# of cbf units       &   128      \\
$\gamma$              &   1.0      \\
$\alpha$              &   0.1      \\
$\lambda_c$           &   1        \\
$\lambda_s$           &   5        \\
$\lambda_f$           &   1        \\
\bottomrule
\end{tabular}
\label{tab:hparams}
\end{table}

\partitle{Deployment and Online Optimization} \\
During deployment, the general aim is to be able to roll out a safe trajectory for as long as possible. At a high level, \method{} is first given a short sequence of state-action pairs as a prompt. Based on the prompt containing the past observations and actions, along with the current observation, the model predicts the next action which is sent to the simulator to get the observation at the next time step. This rollout procedure of generating a new action and the corresponding state is carried on iteratively until a crash happens or a maximum number of timesteps is reached.

At timestep $t$ of the rollout procedure, given the observed history of states and actions represented as the sequence $(s_{t-T+1}, a_{t-T+1}, s_{t-T+2}, a_{t-T+2},...,s_{t-1}, a_{t-1}, s_t)$, we first compute an action proposal for time $t$. Through the online optimization step, our aim is to evaluate whether the proposed action needs to be modified further. We combine the action proposal $\hat{a}_{t}$ with a learnable parameter $\Delta a$ and forward them to the network. Initially, $\Delta a$ is set to zero. We evaluate the violation loss for the CBF condition at time $t$ as:
\begin{equation}
    \mathcal{L}_v=\sigma_+(\eta-C_f(s_t^+,a_t^+)) \\
\end{equation}
where $\eta$ is a parameter representing the safety threshold. We recall that a CBC value of 0 indicates a crash/failure. If $\eta > 0$, that means the CBC value should be greater than a positive value to satisfy the CBF condition, akin to asking the policy to be more conservative. We only perform online optimization if $\mathcal{L}_v$ is nonzero. For this routine, we use the RMSProp optimizer and do $1\sim3$ backward steps to compute $\Delta a$. We present a more detailed ablation study for several online optimization configurations in Appendix~\ref{sec:appendix-ablation}.

\subsection{Baselines}

\partitle{MPC} We implemented an optimization-based model predictive control baseline using the CasAdi solver. At every step, we crop a $5m\times 5m$ neighborhood centered at the MuSHR car from the map, and convert the occupied cells in the neighborhood to circular obstacles. To alleviate the planning burden, we only consider the obstacles that contain both occupied cells and free cells (which means the obstacle is at the boundary of the lidar scan). Then we perform $3\sim 10$ steps of MPC planning to make sure the car is not colliding with any of the obstacles at any time in the MPC horizon. Our best approach involves a 10-step MPC horizon and choose 0.2m as the collision threshold (the car should be at least 0.2m far from the obstacle). After some grid-search, we find that this configuration gives us the best performance.

\partitle{SAC} We trained the Soft-Actor-Critic algorithm~\citep{haarnoja2018soft} for 1000 epochs, where under each epoch, the model rolls out 1000 steps from the simulation and gets 1000 back-propagation updates. The policy net is a 3-layer fully-connected network with hidden units [128, 128, 128] for the layers. The gamma is 0.99, the learning rate is 1e-3, and the batch size is 256.

\partitle{TRPO} We trained the TRPO algorithm~\citep{schulman2015trust} for 1000 epochs, where under each epoch the model rolls out 1000 steps from the simulation and gets 1000 back-prop updates. The policy net is a 3-layer fully-connected network with hidden units [128, 128, 128] for the layers. The discount factor gamma is 0.99, the learning rate is 3e-4, and the batch size is 1000.

\partitle{PPO} We trained the PPO algorithm~\citep{schulman2017proximal} for 1000 epochs, where under each epoch the model rolls out 1000 steps from the simulation and gets 1000 back-prop updates. The policy net is a 3-layer fully-connected network with hidden units [128, 128, 128] for the layers. The discount factor gamma is 0.99, the learning rate is 3e-4, and the batch size is 4000.

For all the RL algorithms, the agent receives a reward of $0.1$ if it does not collide with anything, and $-3$ if it does.

\partitle{GAIL} We trained the GAIL algorithm~\citep{ho2016generative} for 1000 epochs, where we \rbt{only use the safe set of the} expert trajectories that were used in \method{} and we train for 10000 iterations. The GAIL framework consists of a value network, a generator (policy net), and a discriminator. The value network is a 2-hidden-layer fully-connected network with 128 hidden units in each layer and the learning rate is 1e-3. For the generator, we use a 2-hidden-layer fully-connected network with 128 hidden units in each layer. The discount factor gamma for the generator is 0.995 and the learning rate is 3e-4. For the discriminator, we use a 2-hidden-layer fully-connected network with 128 hidden units in each layer, with the learning rate as 1e-4. The sample batch size is 4000.

\partitle{BC} For simple behavior cloning, we trained a 3-hidden-layer fully-connected network with 128 hidden units in each layer for 100 epochs, \rbt{with batchsize 32 and a learning rate of 1e-4, using the same set of safe data used by GAIL.}

\partitle{CPO} For CPO, we adapted the implementation from~\url{https://github.com/SapanaChaudhary/PyTorch-CPO} as the official implementation is coded in Theano, which is incompatible with our simulation framework. We trained the CPO algorithm for 10000 epochs which took approximately 12 hours.

\partitle{SABLAS} For SABLAS, we followed the official implementation~\url{https://github.com/MIT-REALM/sablas} and adapted it to our simulation environment. For CPO and SABLAS, the (safety) constraint is that the shortest lidar beam should be always greater than 0.1m. We trained SABLAS for 20000 iterations over approximately 12 hours. 

\begin{table}[!htbp]
\centering
\caption{Training and deployment time taken}
\begin{tabular}{cccc}
\toprule
Algorithm     & Training time (h/m)   & Runtime (s)      \\ \midrule
MPC & - & 35580 \\
BC & 8h 6m & 579.51 \\
PPO & $>$24h & 560.87 \\
TRPO & $>$24h & 558.62 \\
SAC & $>$24h & 565.24 \\
GAIL & $>$24h & 546.09 \\
PACT & 11h 2m & 666.51 \\
ConBaT & 12h 5m & 852.31 \\
\bottomrule
\end{tabular}
\label{tab:effort}
\end{table}

\section{Additional Results}
\label{sec:appendix-ablation}

\subsection{Computational Effort}

In \Cref{tab:effort}, we outline the computational requirements for the different classes of algorithms we implement in the MuSHR domain. We note MPC to be 1-2 orders of magnitude slower than the learning-based methods during runtime as it requires solving a complex optimization problem at every step, parameterized over the number of obstacles in the neighboring map. While the reinforcement learning baselines are relatively faster in deployment than PACT or \method{}, their training time is much higher. 

\subsection{CBF Critic Ablations}
In \Cref{tab:token_embed}, we compare two different architecture designs for the future state critic $C_f$. CBC-EF (stands for CBC-embedding/future) which takes the current state and action embeddings as input: $\hat{c}_{t+1}=C_f(s_{t}^+, a_{t}^+)$, and CBC-TF (CBC-token/future) which takes the current state embedding but only the current action token as input: $\hat{c}_{t+1}=C_f(s_{t}^+, a'_{t})$. As shown in~\Cref{tab:token_embed}, the online optimization time for CBC-TF is $40\%\sim60\%$ shorter than the runtime of CBC-EF, which is because in CBC-TF the action token does not have to be processed through the Transformer and temporally-fused with features from other timesteps.
However, we find that consistently among all the three tracks, CBC-EF achieves much better collision rate and ATL compared to CBC-TF. We attribute to the feature richness of the action embeddings after fusing with other state-action history, compared to the raw action tokens. Thus, we primarily use the CBC-EF architecture in our paper.

\begin{table}[!htbp]
\centering
\vspace{-1mm}
\centering
\caption{Using action embedding vs. action token for the Control Barrier Critic}
\resizebox{0.45\textwidth}{!}{%
\begin{tabular}{cc|c|c|c|c|c}
    \toprule
    \multirow{2}{*}{Track} &
    \multicolumn{2}{c|}{Collision Rate (\%)} & \multicolumn{2}{c|}{ATL (\# steps)} & \multicolumn{2}{c}{Runtime (sec)}\\
    \cline{2-7} \addlinespace
     & CBC-EF & CBC-TF & CBC-EF & CBC-TF & CBC-EF & CBC-TF \\ \midrule
Playground    & 0  &  1.5  & 1000 & 983.46  & 134.17 & 76.78 \\
Silverstone   & 0 & 54.6 & 1000 & 632.28    & 147.61 & 81.94 \\
Austin   & 61.7 & 96.8 & 678.15   & 279.13  & 180.88 & 76.62 \\
    \bottomrule
\end{tabular}
}
\label{tab:token_embed}
\end{table}

\subsection{Online Optimization Ablations}

\Cref{tab:optim_ablations} contains the results from several ablation studies we perform on the F1/10 dataset to identify how the online optimization routine behaves under varying influence of its hyperparameters. We examine the effects of these three parameters: a) the number of gradient descent steps, b) the learning rate, and c) the CBF safety threshold $\eta$. From \Cref{tab:optim_ablations}(a), we note that increasing the optimization steps in the online optimization does not improve the performance. From \Cref{tab:optim_ablations}(b), we see that a fairly small learning rate can already make the online optimization achieve collision-free performance on F1/10, whereas a larger learning rate leads to a more unstable optimization process hence deteriorating the result. From \Cref{tab:optim_ablations}(c), we see that the threshold $\eta$ to some degree reflects the conservativeness - a larger threshold will result in a more conservative policy, which potentially can lead to better performance.

\begin{table}[ht]
\small
\centering
\begin{subtable}{0.25\textwidth}
    \centering
    \caption{SGD step.}
    \begin{tabular}{ccc}
    \toprule
                Steps & Collision & ATL   
                \\ \midrule
    0   & 1	&175.45
  \\
    1   & 0.0078	& 991.24
    \\
    2  & 0.0078 &	991.22
  \\
    3        & 0.0078	&991.21
 \\
    4    & 0.0078	&991.21
 \\
    \bottomrule
    \end{tabular}

\end{subtable}
\qquad
\begin{subtable}{0.25\textwidth}
    \centering
    \caption{Learning rate (lr)} 
    \begin{tabular}{ccc}
    \toprule
               lr & Collision & ATL      
               \\ \midrule

    0.05 & 0 &	1000	
 \\
    0.1  & 0.0078&	991.23	
 \\
    0.2   & 0.0078 &	991.22	
 \\
    0.3   & 0.0156	&983.4
\\
    0.5  & 0.0234 & 	975.73
 \\
    \bottomrule
    \end{tabular}
       
\end{subtable}
\qquad
\begin{subtable}{0.25\textwidth}
    \centering
    \caption{Threshold} 
    \begin{tabular}{ccc}
    \toprule
               Threshold & Collision & ATL      
               \\ \midrule
    -0.1 & 0.875 &	343.6
 \\
    -0.05  & 0.3672&714.08
 \\
    0.0   & 0.0469&	958.29
 \\
    0.05   & 0.0156	&983.47
\\
    0.2   & 0.0078	&991.22
\\
    \bottomrule
    \end{tabular}
       
\end{subtable}
\caption{Ablation studies for online optimization (iterations, learning rates and thresholds)}
\label{tab:optim_ablations}
\end{table}

\subsection{Learned Control Barrier Critic Visualization}

\begin{figure}[!htbp]
\centering
\begin{subfigure}{0.48\textwidth}
\centering
\includegraphics[width=0.95\textwidth]{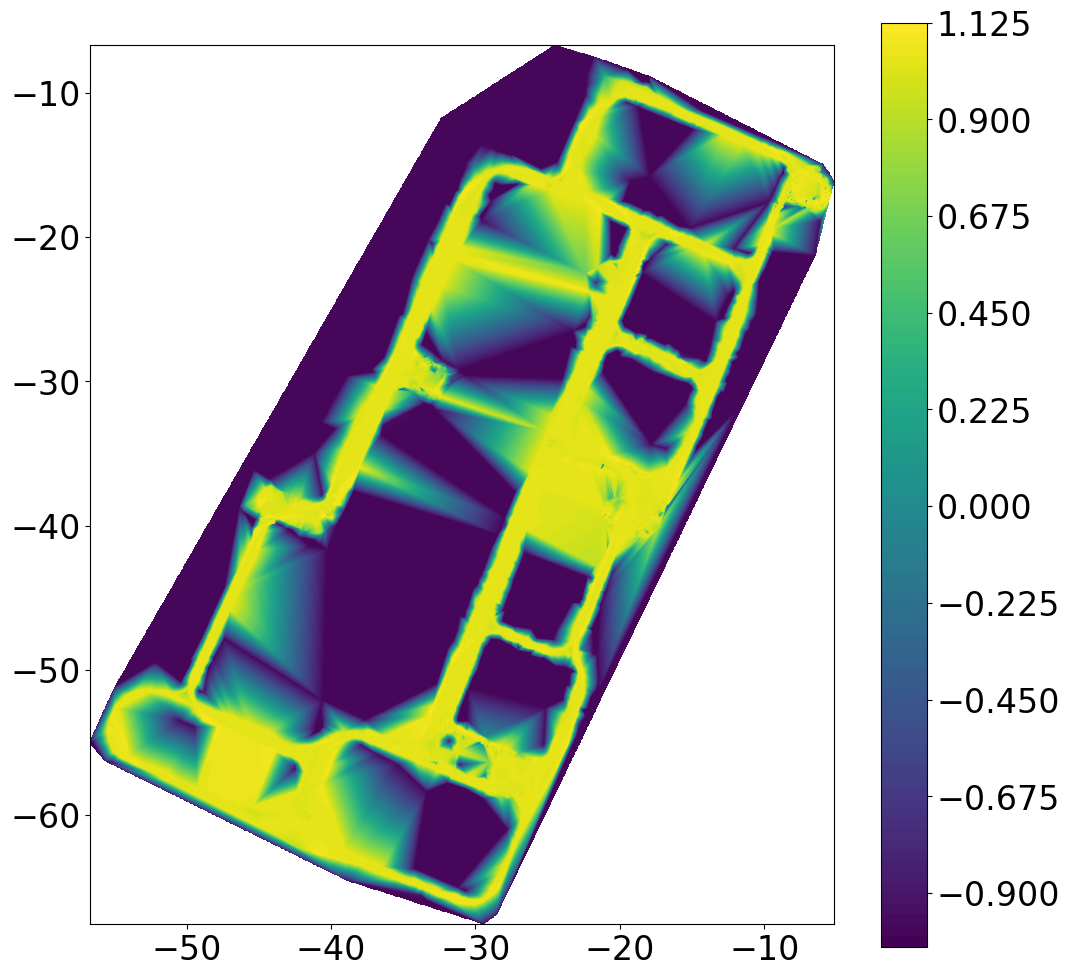}
    \caption{Visualization from Control Barrier Critics (CBC)}
    \label{fig:cbf_map}
\end{subfigure}
\hfill
\begin{subfigure}{0.48\textwidth}
\centering
    \includegraphics[width=0.8\textwidth]{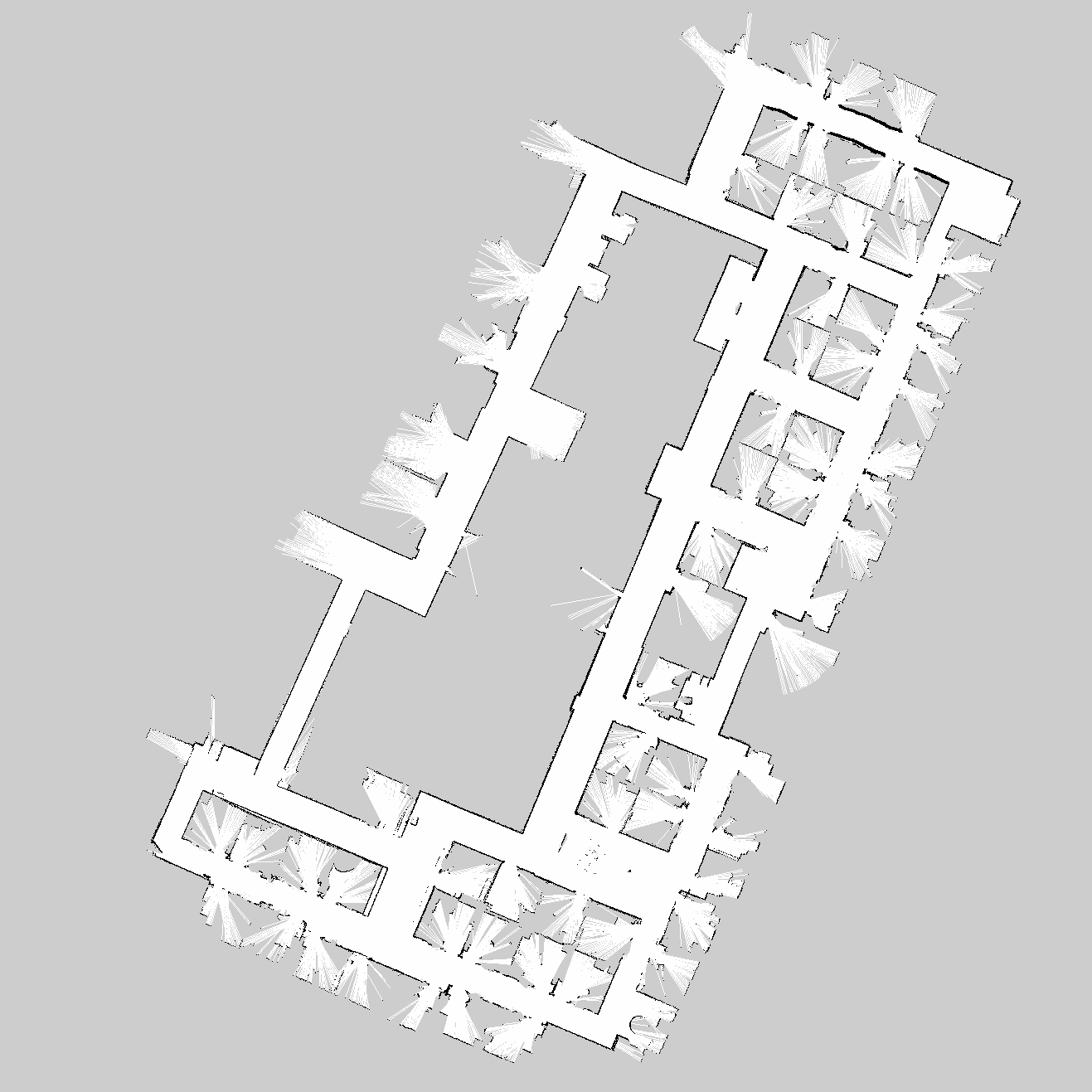}
    \caption{Visualization of the original MuSHR environment}
    \label{fig:mushr_occ}
\end{subfigure}
\caption{Visualization of Control Barrier Critics for MuSHR}
\label{fig:cbf_occ_compare}
\end{figure}

To inspect how the control barrier critic (CBC) is learned, we plot the CBC prediction along the expert trajectories and interpolate over unvisited regions on the map. As shown in Figure~\ref{fig:cbf_map}, the CBC is able to predict safe values in the center of the pathways, and predict negative values in regions that are close to the boundary of the wall, which is consistent with the expectation that the closer to the center of the hallway, the safer. However, we do notice that there exists unsafe area that the CBC mistakenly marks that area as ``safe" (e.g., around the location (-30, -20)), which could be due to interpolation error or network inability to predict the safety score at that spot.

\rbt{\subsection{Trade-off between safety and optimality -MuSHR domain}}

\rbt{To investigate how \method{} will affect the optimality of the solution, we design a goal-reaching task under the MuSHR simulation environment, collect expert demonstrations and train PACT and \method{} to reach the destination point. Specifically, the task is to reach a fixed destination point $(-9.2, -17.5)^T$ on the map from a randomly initialized position. We collect 10000 expert trajectories using the search-based MPC, which is the same one used in the previous MuSHR data collection process in the main paper, but with a fixed destination point (and 10000 different initialized points).} 

\rbt{We train the PACT and \method{} on the expert data, following the same set of hyperparameters used in the main paper. During testing, we test for 128 different initial starting points, use the controller trained by PACT/\method{} to rollout for at most 5000 time steps for evaluation. We define the success rate as the percentage of rollout trajectories that can reach the goal without any collision, compute the average trajectory length before reaching goal, and the average trajectory length before collision.}

\begin{figure}[!htbp]
\centering
\includegraphics[width=0.48\textwidth]{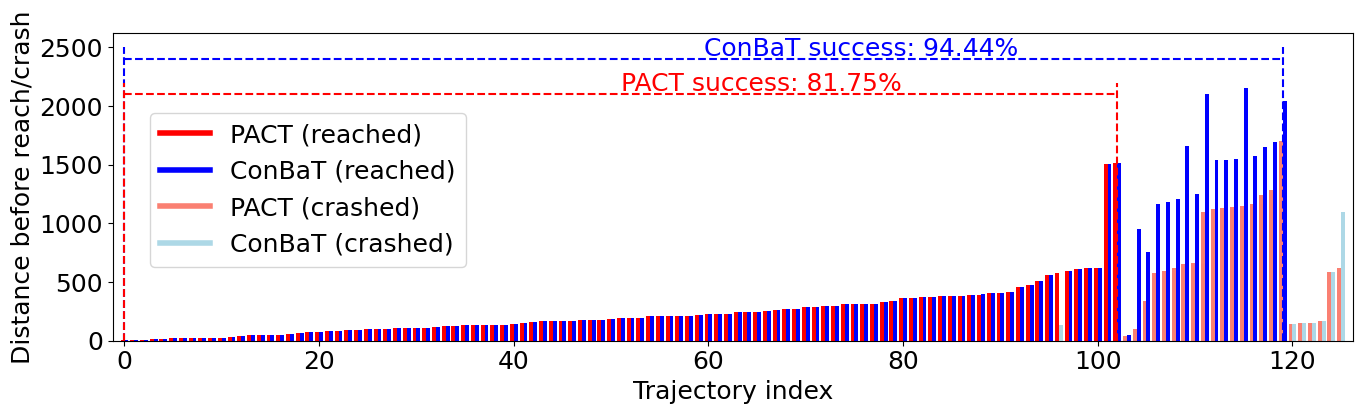}
\caption{\small{\rbt{Trajectory comparison between PACT and \method{} for the MuSHR goal-reaching task. \method{} results in higher goal-reaching success rate, and can preserve the same solution quality as the PACT model when the PACT trajectory is not crashed. \method{} also results in 64\% overhead for the average goal-reaching distance, which happens when the PACT trajectories crashes.}}}
\label{fig:optimality_test}
\end{figure}

\rbt{We plot the length of each trajectory before goal-reaching/collision for each method for comparison in Figure~\ref{fig:optimality_test}. We categorize the trajectories depending on the "goal-reaching/crashing" consequence of the PACT/\method{} trajectories, and inside each region we sort those trajectories based on the PACT trajectory length before crashing/goal-reaching. As shown in Figure~\ref{fig:optimality_test}, \method{} achieves a success rate of 94.44\%, which is 12.69\% higher than the PACT model. Our method doesn't improve the optimality/quality of each individual solution, because \method{} is designed for improving safety rather than optimality.
However, compared to PACT model, \method{} maintains the quality of the solution (indices 0-102) when the PACT trajectories are safe and only increases the trajectory length when the PACT trajectories are crashing (indices 103-127). This shows \method{} can work in a shield-like fashion which preserves the base model behavior when safe, and only changes the trajectory when the potential unsafe case emerges.}

\rbt{\subsection{Ablation study for CBF trend loss}}
\rbt{One might think it is straight-forward to use safe/unsafe labels to guide the safe learning process. Here we emphasize the importance of the smoothness loss (trend loss). From the same PACT base model under MuSHR environment, we train the ConBaT to learn the safety score by differnet weighting for the smoothness term (ranging from 0-50, where in the main paper we use $\mathcal{L}_s=5$). We follow the same optimization process and rollout mechanism in rollout. As shown in Table~\ref{tab:abl_smoothness}, without the smoothness loss or when $\lambda_s\leq 2.0$, the \method{} cannot improve upon the PACT performance. This makes sense as a smaller trend is not advance enough to prevent the states fall into unsafe region (but too large trend loss will hurt the CBF classification result hence affect the correctness for the CBF-critic to judge whether next state is really safe or unsafe). With large smoothness loss ($\lambda_s\geq 5.0$) added to the \method{} training, the performance gets better than PACT and the peak is from $\lambda_s=5$. Thus we picked $\lambda_s=5$ in our experiments.}

\begin{table}[!htbp]
\centering
\begin{tabular}{cccc}
\toprule
     Method & Smoothness weight $\lambda_s$   &  Collision (\%) & ATL (\# steps) \\ \midrule
PACT & - & 0.35 & 3453.34\\
\method{} & 0 & 0.63 & 2799.79\\
\method{} & 0.1 & 0.88 & 2086.95 \\
\method{} & 0.2 & 0.88 & 2004.23\\
\method{} & 0.5 & 0.76 & 2262.83\\
\method{} & 1 & 0.66 & 2991.55\\
\method{} & 2 & 0.91 & 2147.82\\
\textbf{\method{}} & \textbf{5} & \textbf{0.18} & \textbf{4271.54}\\
\method{} & 10 & 0.23 & 3944.15\\
\method{} & 20 & 0.30 & 3654.66\\
\method{} & 50 & 0.35 & 3453.34\\
\bottomrule
\end{tabular}
\caption{\rbt{\method{} training under different smoothness weights.}}
\label{tab:abl_smoothness}
\end{table}

\subsection{Different architectures for safe policy learning on MuSHR} 

\begin{table}[!htbp]
\centering
\setlength\tabcolsep{3.5pt}
\begin{tabular}{cccccc}
\toprule
     Architecture & Params. & Training & Runtime (s)&  Collision (\%) & ATL \\ \midrule
MLP  & \textbf{243k} & \textbf{8h 43m} & \textbf{351.47} & 34.38 & 3580.01\\
MLP+CBC& 293k & 9h 47m & 419.42 & 28.91 & 3721.06 \\
LSTM  & 343k & 9h 49m & 383.94 & 53.12 & 2652.30\\
LSTM+CBC & 393k & 10h 54m & 425.09 & 47.66 & 2802.03 \\
GRU   & 376k & 8h 51m & 373.15 & 35.16 & 3468.57\\
GRU+CBC & 426k & 9h 56m & 382.02 & 31.25 & 3613.84\\
PACT   & 310k & 11h 2m & 383.32& 35.16 & 3453.34\\
\textbf{PACT+CBC} & 360k & 12h 5m & 411.34 & \textbf{17.97} & \textbf{4271.54}\\
\bottomrule
\end{tabular}
\caption{Policy learning on MuSHR dataset under varied architectures and optimization options (CBC). The runtime is different from Table~\ref{tab:effort} because of the different GPU resources used here.}
\label{tab:abl_arch}
\end{table}

To study the importance of the Transformer backbone in safe policy learning, we replace it with different architectures and repeat the same training, fine-tuning, and test procedures as the \method{}. To be more specific, we consider three other architectures, Multi-layer Perception (MLP), Long Short-Term Memory (LSTM), and Gated Recurrent Units (GRU). For MLP, for each time step, we simply concatenate the current input tokens with all its history input tokens (pad with zero if none) with the 16-step time horizon and send them to the MLP backbone (2 hidden layers with 128 neurons in each layer) to generate the output embeddings. For LSTM and GRU, we use two recurrent layers with a hidden state size of 128; we have a projection layer to cast the final output (which has the same dimension as the hidden states) to the output embedding space. For each backbone, we record the performance of the model trained in the first stage (without CBC learning or optimization, denote as ``Backbone Name") as well as after the second stage (with CBC learning and optimization, similar to \method{}, denote as ``Backbone Name+CBC."). To make a fair comparison, we do a grid search for each architecture over 50 configurations for the online optimization phase and report the best one. As shown in Table~\ref{tab:abl_arch}, in the first stage, the MLP backbone achieves the lowest collision rate and the highest ATL, slightly better than GRU and transformer backbones. And in the second stage, all the backbones can achieve $4\%\sim17\%$ collision rate improvement after the optimization process, which indicates that \method{} is architecture-agnostic. Among all the architectures, the Transformer backbone achieves the most significant reduction in the collision rate ($17\%$) and achieves the lowest collision rate. This is likely because the safety-ness in MuSHR depends on both the current state and the history state/action information, and the Transformer backbone excels at modeling the long-term dependencies. Hence the CBC can better tell the safety scores of the states, and correspondingly, the controller can achieve the best performance.

\end{document}